%% file: PaperForReview.tex
\documentclass[10pt,twocolumn,letterpaper]{article}
\usepackage[pagenumbers]{cvpr} %
\usepackage{graphicx}
\usepackage{amsmath}
\usepackage{amssymb}
\usepackage{booktabs}
\usepackage{colortbl}
\usepackage{multirow}
\usepackage{makecell, tabularx}

\usepackage[pagebackref,breaklinks,colorlinks]{hyperref}
\usepackage[capitalize]{cleveref}

\crefname{section}{Sec.}{Secs.}
\Crefname{section}{Section}{Sections}
\Crefname{table}{Table}{Tables}
\crefname{table}{Tab.}{Tabs.}

\def\cvprPaperID{} %
\def\confName{CVPR}
\def\confYear{2023}

\begin{document}

\title{InstantAvatar: Learning Avatars from Monocular Video in 60 Seconds}

\author{Tianjian Jiang$^{1}$*, Xu Chen$^{1,2}$*, Jie Song$^{1}$ , Otmar Hilliges$^{1}$ \\
$^{1}$ ETH Zürich \quad $^{2}$ Max Planck Institute for Intelligent Systems, T{\"u}bingen
\\
\vspace{.1cm}\href{https://tijiang13.github.io/InstantAvatar}{https://tijiang13.github.io/InstantAvatar/}
}

\twocolumn[{
\maketitle
\vspace*{-7ex}
\begin{center}
\includegraphics[width=\textwidth]{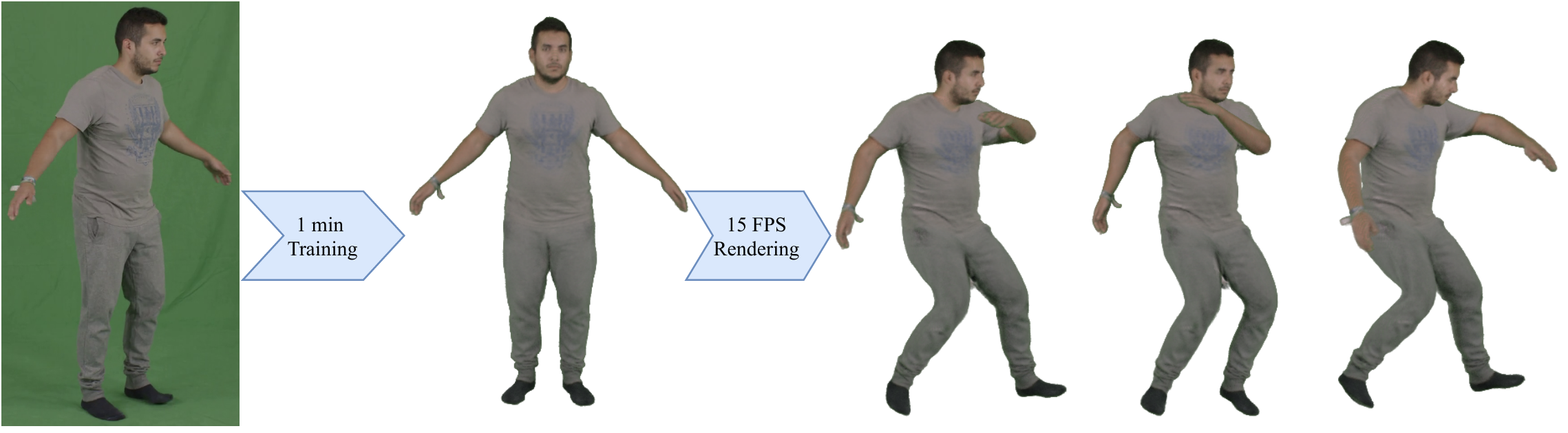}
\end{center}
\vspace*{-10px}
\captionof{figure}{\textbf{InstantAvatar:} we propose a system that can reconstruct animatable high-fidelity human avatars from monocular video within 60 seconds, providing poses and masks, and can animate and render the model at 15 FPS at $540\times540$ resolution. To achieve this we integrate accelerated neural radiance fields, originally designed for rigid scenes, with a fast correspondence search module for articulation. An efficient empty-space skipping strategy further speeds up training and inference, enabling  near-instant avatar learning. }
\label{fig:teaser}
}
\vspace{20px}
]

\def\thefootnote{*}\footnotetext{Equal Contribution.}

\input{notations}
\input{sec/01-abstract}
\input{sec/02-intro}
\input{sec/03-related_work}

\input{sec/04-method}
\input{sec/05-experiment}
\input{sec/07-conclusion}

\clearpage
{\small
\bibliographystyle{ieee_fullname}
\bibliography{bibliography_long,bibliography,bibliography_custom}

}

\end{document}

%% file: notations.tex
\newcommand{\netparam}{\sigma}
\newcommand{\point}{\mathbf{x}}
\newcommand{\spoint}{\hat{\mathbf{x}}}
\newcommand{\function}[1]{{#1}_{\netparam_{#1}}}
\newcommand{\mfunction}[1]{\mathbf{#1}_{\netparam_{#1}}}
\newcommand{\raydir}{\mathbf{v}}
\newcommand{\rayori}{\mathbf{c}}

\newcommand{\pixel}{\mathbf{p}}
\newcommand{\occ}{o}
\newcommand{\loss}{\mathcal{L}}
\newcommand{\bodypose}{\boldsymbol{\theta}}
\newcommand{\bone}{\boldsymbol{B}}
\newcommand{\nbone}{n_{\boldsymbol{B}}}

\newcommand{\texture}{\mathbf{t}}

\newcommand{\gt}{}
\newcommand{\jac}{\mathbf{J}}

\newcommand{\name}{Fast-SNARF\xspace}
\newcommand{\snarf}{SNARF\xspace}
\newcommand{\suppmat}{supp. mat.\xspace}

\newcommand{\methodname}{InstantAvatar\xspace}

%% file: sec/01-abstract.tex
\begin{abstract}
In this paper, we take a significant step towards real-world applicability of monocular neural avatar reconstruction by contributing \methodname, a system that can reconstruct human avatars from a monocular video within seconds, and these avatars can be animated and rendered at an interactive rate. 
To achieve this efficiency we propose a carefully designed and engineered system, that leverages emerging acceleration structures for neural fields, in combination with an efficient empty space-skipping strategy for dynamic scenes. We also contribute an efficient implementation that we will make available for research purposes. 
Compared to existing methods, \methodname converges 130$\times$ faster and can be trained in minutes instead of hours. It achieves comparable or even better reconstruction quality and novel pose synthesis results. When given the same time budget, our method significantly outperforms SoTA methods. \methodname can yield acceptable visual quality in as little as 10 seconds training time. 
\end{abstract}

%% file: sec/02-intro.tex
\section{Introduction}
\label{sec:intro}
Creating high-fidelity digital humans is important for many applications including immersive tele-presence, AR/VR, 3D graphics, and the emerging metaverse. 
Currently acquiring personalized avatars is an involved process that typically requires the use of calibrated multi-camera systems and incurs significant computational cost. In this paper, we embark on the quest to build a system for the learning of 3D virtual humans from monocular video alone that is lightweight enough to be widely deployable and fast enough to allow for walk-up and use scenarios.

The emergence of powerful neural fields has enabled a number of methods for the reconstruction of animatable avatars from monocular videos of moving humans~\cite{alldieck2018video,Alldieck20183DV,chen2021animatable,Peng2021CVPR,weng2022humannerf}. 
These methods typically model human shape and appearance in a pose-independent canonical space.
To reconstruct the model from images that depict humans in different poses, such methods must use animation (e.g. skinning) and rendering algorithms, to deform and render the model into posed space in a differentiable way. This mapping between posed and canonical space allows optimization of network weights by minimizing the difference between the generated pixel values and real images.
Especially methods that leverage neural radiance fields (NeRFs)~\cite{mildenhall2020nerf} as the canonical model have demonstrated high-fidelity avatar reconstruction results. However, due to the dual need for differentiable deformation modules and for volume rendering, these models require hours of training time and cannot be rendered at interactive rates, prohibiting their broader application.

In this paper, we aim to take a significant step towards real-world applicability of monocular neural avatar reconstruction by contributing a method that takes no longer for reconstruction, than it takes to capture the input video. 
To this end, we propose \methodname a system that reconstructs high-fidelity avatars within 60 seconds, instead of hours, given a monocular video, pose parameters and masks. Once learned the avatar can be animated and rendered at interactive rates.
Achieving such a speed-up is clearly a challenging task that requires careful method design, requires fast differentiable algorithms for rendering and articulation, and requires efficient implementation. 

Our simple yet highly efficient pipeline combines several key components. 
First, to learn the canonical shape and appearance we leverage a recently proposed efficient neural radiance field variant~\cite{mueller2022instant}. Instant-NGP accelerates neural volume rendering by replacing multi-layer perceptrons (MLP) with  a more efficient hash table as data structure. However, because the spatial features are represented explicitly, Instant-NGP is limited to rigid objects. 
Second, to enable learning from posed observations and to be able to animate the avatar, we interface the canonical NeRF with an efficient articulation module, Fast-SNARF~\cite{fastsnarf}, which efficiently derives a continuous deformation field to warp the canonical radiance field into the posed space. Fast-SNARF is orders of magnitude faster compared to its much slower predecessor~\cite{chen2021snarf}. 

Finally, simply integrating existing acceleration techniques is not sufficient to yield the desired efficiency (see Tab.~\ref{tab:abltion_speed}). With acceleration structures for the canonical space and a fast articulation module in place, rendering the actual volume becomes the computational bottleneck. To compute the color of a pixel, standard volume rendering needs to query and accumulate densities of hundreds of points along the ray. A common approach to accelerating this is to maintain an occupancy grid to skip samples in the empty space. However, such an approach assumes rigid scenes and cannot be applied to dynamic scenes such as humans in motion.

We propose an empty space skipping scheme that is designed for dynamic scenes with known articulation patterns. At inference time, for each input body pose, we sample points on a regular grid in posed space and map them back to the canonical model to query densities. Thresholding these densities yields an occupancy grid in canonical space, which can then be used to skip empty space during volume rendering. For training, we maintain a shared occupancy grid over all training frames, recording the union of occupied regions over individual frames. This occupancy grid is updated every few training iterations with the densities of randomly sampled points, in the posed space of randomly sampled frames. This scheme balances computational efficiency and rendering quality.

We evaluate our method on both synthetic and real monocular videos of moving humans, and compare it with state-of-the-art methods on monocular avatar reconstruction. Our method achieves on-par reconstruction quality and better animation quality in comparison to SoTA methods, while only requiring minutes of training time instead of more than 10 hours.
When given the same time budget, our method significantly outperforms SoTA methods. We also provide an ablation study to demonstrate the effect of our system's components on speed and accuracy.

%% file: sec/03-related_work.tex
\section{Related Work}
\label{sec:related_work}
\paragraph{3D Human Reconstruction} Reconstructing 3D human appearance and shape is a long-standing problem. High-quality reconstruction has been achieved in ~\cite{matusik2000image,collet2015high,dou2016fusion4d,guo2019relightables} by fusing observations from a dense array of cameras or depth sensors. The expensive hardware requirement limits such methods to professional settings. Recent work~\cite{alldieck2018video,Alldieck20183DV,deepcap,Xu:2018:MHP:3191713.3181973,habermann2019livecap,guo2021human,jiang2022hifecap} demonstrates 3D human reconstruction from a monocular video by leveraging personalized or generic template mesh models such as SMPL~\cite{Loper2015SIGGRAPH}. These methods reconstruct 3D humans by deforming the template to fit 2D joints and silhouettes. However, personalized template mesh might not be available in many scenarios and generic template mesh cannot model high-fidelity details and different clothing topologies. 

Recently, neural representations~\cite{Park2019CVPRDeepsdf,Mescheder2019CVPR,Oechsle2019ICCV,Mildenhall2020ECCV} have emerged as a powerful tool to model 3D humans~\cite{Chibane2020CVPR,guo2021human,Saito2019ICCV, Saito2020CVPR,xiu2022icon,zheng2021pamir,Mihajlovic2021CVPR, wang2021metaavatar, he2020geopifu,chen2022gdna, hong2022avatarclip, noguchi2022unsupervised, bergman2022gnarf, zhang2022avatargen,Deng2020ECCV,chen2021snarf,Tiwari2021ICCVNeuralGIF, lin2022fite,dong2022pina,Huang2020CVPR,he2021arch++,Corona2021CVPRSMPLicit,Palafox2021ICCV,xiu2022icon,Peng2021CVPR, peng2021animatable,2021narf,liu2021neural,chen2021animatable, weng2022humannerf,jiang2022selfrecon,li2022tava,ARAH:2022:ECCV,jiang2022neuman,xu2021h,Feng2022scarf}. Using neural representations, many works~\cite{Peng2021CVPR, peng2021animatable,2021narf,liu2021neural,chen2021animatable, weng2022humannerf,jiang2022selfrecon,li2022tava,ARAH:2022:ECCV,jiang2022neuman,xu2021h} can directly reconstruct high fidelity neural human avatars from a sparse set of views or a monocular video without pre-scanning personalized template. These methods model 3D human shape and appearance via neural radiance field~\cite{mildenhall2020nerf} or signed distance and texture field in a pose-independent canonical space and then deform and render the model into various body poses in order to learn from posed observations. While achieving impressive quality and can learn avatars from a monocular video, these methods suffer from slow training and rendering speed due to the slow speed of the canonical representation as well as deformation algorithms. Our method addresses this issue and enables learning avatars within minutes.

\paragraph{Accelerating Neural Radiance Field}
Several methods have been proposed to improve the training and inference speed of neural representations~\cite{Liu20neurips_sparse_nerf, takikawa2021nglod, SunSC22, mueller2022instant, Chen2022ECCV, liu2020nsvf,yu2021plenoctrees,yu2022plenoxels,garbin2021fastnerf,reiser2021kilonerf,li2022nerfacc}.
The core idea is to replace MLPs in neural representations with more efficient representations. A few works~\cite{liu2020nsvf,yu2022plenoxels,yu2021plenoctrees} propose to use voxel grids to represent neural fields and achieve fast training and inference speed.
Instant-NGP~\cite{mueller2022instant} further replaces dense voxels with a multi-resolution hash table, which is more memory efficient and hence can record high-frequency details. Besides improving the efficiency of the representation, several works~\cite{mueller2022instant,li2022nerfacc,Liu20neurips_sparse_nerf} also improve the rendering efficiency by skipping empty space via an occupancy grid to further increase training and inference speed.

While achieving impressive quality and training efficiency, these methods are specifically designed for rigid objects. Generalizing these methods to non-rigid objects is not straightforward. We combine Instant-NGP with a recent articulation algorithm to enable animation and learning from posed observations. In addition, we propose an empty space skinning scheme for dynamic articulated humans.

%% file: sec/04-method.tex
\begin{figure*}[ht]
\begin{center}
    \includegraphics[width=\linewidth]{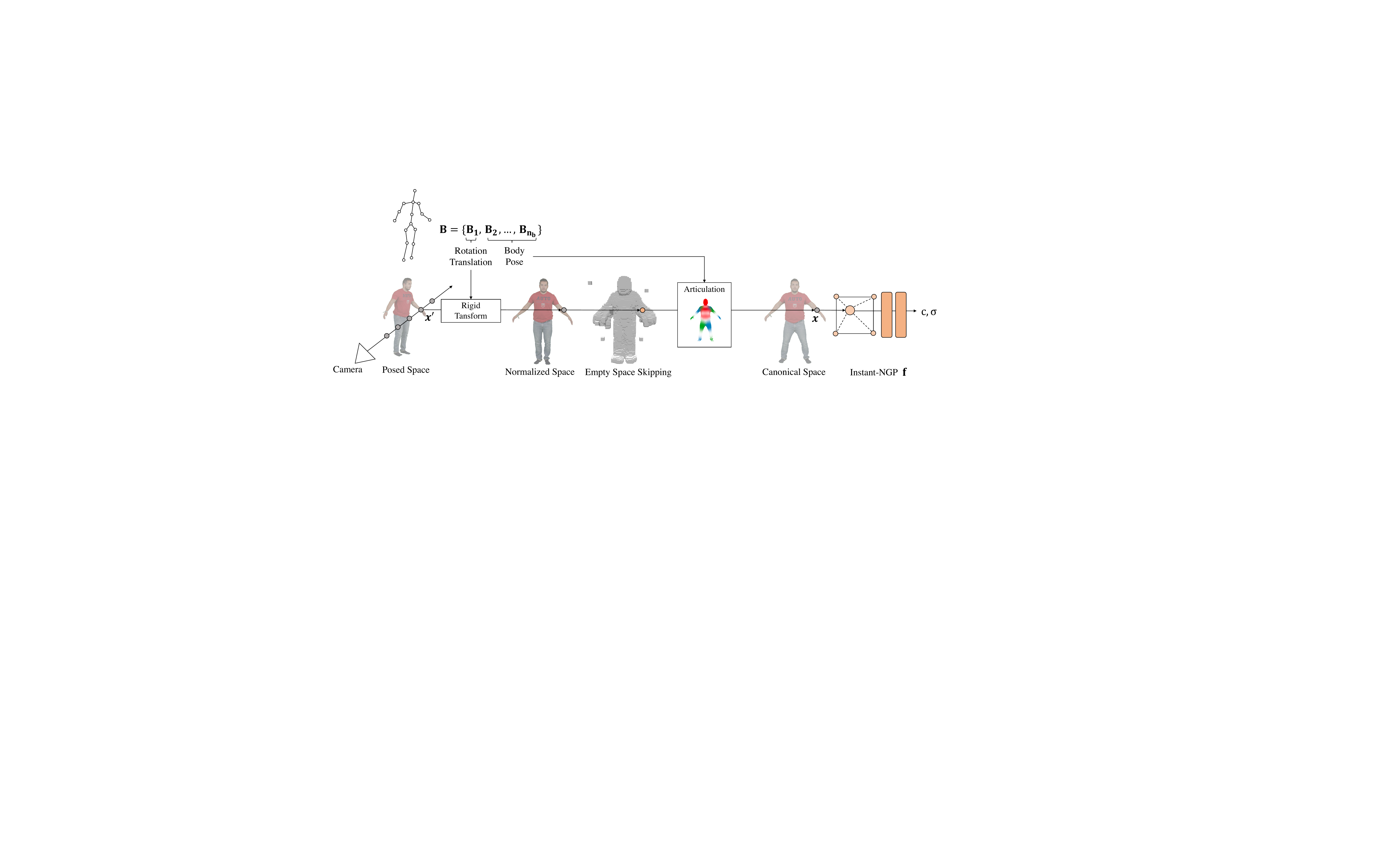}
    \caption{\textbf{Method Overview.} For each frame, we sample points along the rays in posed space. We then transform these points into a normalized space where we remove the global orientation and translation of the person. In this normalized space, we filter points in empty space using our occupancy grid. The remaining points are deformed to canonical space using an articulation module and then fed into the canonical neural radiance field to evaluate the color and density. 
    \vspace{-2em}
    }
    \label{fig:overview}
\end{center}
\end{figure*}

\section{Method}
\label{sec:pipeline}
Given a monocular video of a moving human, our primary goal is to reconstruct a 3D human avatar within a tight computational budget. In this section, we first describe the preliminaries that our method is based on (Sec.~\ref{sec:method:preliminaries}), which include an accelerated neural radiance field that we use to model the appearance and shape in canonical space and an efficient articulation module to deform the canonical radiance field into posed space.
We then describe our implementation of the volumetric renderer to produce images from the radiance fields in an efficient manner (Sec.~\ref{sec:method:volume rendering}). 
To avoid inefficient sampling of empty space, we leverage the observation that the 3D bounding box around the human body is dominated by empty space. We then propose an empty space skipping scheme specifically designed for humans (Sec.~\ref{sec:method:accelerating}). 
Finally, we discuss training objectives and regularization strategies (Sec.~\ref{sec:method:losses}).

\subsection{Preliminaries}
\label{sec:method:preliminaries}
\paragraph{Efficient Canonical Neural Radiance Field}
We model human shape and appearance in a canonical space using a radiance field $\mfunction{f}$, which predicts the density $\sigma$ and color $c$ of each 3D point $\point$ in the canonical space:
\begin{align}
\mfunction{f}: \mathbb{R}^3 &\rightarrow \mathbb{R}^+, \mathbb{R}^3 \\
\point &\mapsto \sigma,c
\end{align}
where $\sigma_{f}$ are the parameters of the radiance field.

We use Instant-NGP~\cite{mueller2022instant} to parameterize $\mfunction{f}$, which achieves fast training and inference speed by using a hash table to store feature grids at different coarseness scales. To predict the texture and geometry properties of a query point in space, they read and tri-linearly interpolate the features at its neighboring grid points and then concatenate the interpolated features at different levels. The concatenated features are finally decoded with a shallow MLP. 

\paragraph{Articulating Radiance Fields}
\label{sec:method:canonical_model}
To create animations and to learn from posed images, we need to generate deformed radiance fields in target poses $\mfunction{f}'$. The posed radiance field is defined as 
\begin{align} 
\mfunction{f}': \mathbb{R}^3 &\rightarrow \mathbb{R}^+, \mathbb{R}^3 \\
\point' &\mapsto \sigma,c ,
\end{align}
which outputs color and density for each point in posed space. 
We use a skinning weight field $\mathbf{w}$ in canonical space to model articulation, with $\sigma_{w}$ being its parameters:
\begin{align} 
\mfunction{w}: \mathbb{R}^3 &\rightarrow \mathbb{R}^{n_b} , \\
\point &\mapsto w_1, ..., w_{n_b} .
\end{align}
where $n_b$ is the number of bones in the skeleton. To avoid the computational cost of \cite{chen2021snarf}, \cite{fastsnarf} represents this skinning weight field as a low-resolution voxel grid. 
The value of each grid point is determined as the skinning weights of its nearest vertex on the SMPL~\cite{Loper2015SIGGRAPH} model. With this the canonical skinning weight field and target bone transformations $\mathbf{B} = \{\mathbf{B}_1, ..., \mathbf{B}_{n_b}\}$, a point $\point$ in canonical space is transformed to deformed space $\point'$ via linear blend skinning as following:
\begin{align}
    \point' &= \textstyle{\sum_{i=1}^{n_b}} w_i \mathbf{B}_i \point
    \label{eqn:lbs}
\end{align}
The canonical correspondences $\point^*$ of a deformed point $\point'$ are defined by the inverse mapping of Equation.~\ref{eqn:lbs}. The key is to establish the mapping from points in posed space $\point'$ to their correspondences in the canonical space $\point^*$. This is efficiently derived by root-finding in Fast-SNARF~\cite{fastsnarf}. The posed radiance field $\mfunction{f}'$ can then be determined as $\mfunction{f}'(\point') = \mfunction{f}(\point^*)$.

\subsection{Rendering Radiance Fields}
\label{sec:method:volume rendering}

The articulated radiance field $\mfunction{f}'$ can be rendered into novel views via volume rendering. Given a pixel, we cast a ray $\mathbf{r} = \mathbf{o} + t \mathbf{d}$ with $\mathbf{o}$ being the camera center and $\mathbf{d}$ being the ray direction. We sample $N$ points $\{\point'_i\}^N$ along the ray between the near and far bound, and query the color and density of each point from the articulated radiance field $\mfunction{f}'$ by mapping $\{\point'_i\}^N$ back to the canonical space and querying from the canonical NeRF model $\mfunction{f}$, as illustrate in Fig.~\ref{fig:overview}. We then accumulate queried radiance and density along the ray to get the pixel color $C$
\begin{align}
C = \sum_{i=1}^N \alpha_i \prod_{j < i} (1 - \alpha_j) c_i ,
\text{with \ } \alpha_i = 1 - \exp(\sigma_i \delta_i)
\end{align}
where $\delta_i = \|\point'_{i+1} - \point'_i\|$ is the distance between samples.

While the acceleration modules of Sec.~\ref{sec:method:preliminaries} already achieve significant speed-up over the vanilla variants (NeRF~\cite{mildenhall2020nerf}, SNARF~\cite{chen2021snarf}), the rendering itself now becomes the bottleneck. 
In this paper, we optimize the process of neural rendering, specifically for the use-case of dynamic humans.

\subsection{Empty Space Skipping for Dynamic Objects}
\label{sec:method:accelerating}
We note that the 3D bounding box surrounding the human body is dominated by empty space due to the articulated structure of 3D human limbs. This results in a large amount of redundant sample queries during rendering and hence significantly slows down rendering. For rigid objects, this problem is eliminated by caching a coarse occupancy grid and skipping samples within non-occupied grid cells. However, for dynamic objects, the  exact location of empty space varies across different frames, depending on the pose.  

\paragraph{Inference Stage}
At inference time, for each input body pose, we sample points on a $64\times64\times64$ grid in posed space and query their densities from the posed radiance field $\mfunction{f}'$. We then threshold these densities into binary occupancy values. To remove cells that have been falsely labeled as empty, due to the low spatial resolution, we dilate the occupied region to fully cover the subject. Due to the low resolution of this grid and the large amount of queries required to render an image, the overhead to construct such an occupancy grid is negligible.

During volumetric rendering, for point samples inside the non-occupied cells, we directly set their density to zero without querying the posed radiance field $\mfunction{f}'$. This reduces unnecessary computation to a minimum and hence improves the inference speed.

\paragraph{Training Stage} During training, however, the overhead to construct such an occupancy grid at each training iteration is no longer negligible. To avoid this overhead, we construct a \emph{single} occupancy grid for the entire sequence by recording the union of occupied regions in each of the individual frames. Specifically, we build an occupancy grid at the start of training and update it every $k$ iterations, by taking the moving average of the current occupancy values and the densities queried from the posed radiance field $\mfunction{f}'$ at the current iteration. Note that this occupancy grid is defined in a normalized space where the global orientation and translation are factored out so that the union of the occupied space is as tight as possible and hence unnecessary queries are further reduced.

\subsection{Training Losses}
\label{sec:method:losses}
We train our model by minimizing the robust Huber loss $\rho$ between the predicted color of the pixels $C$ and the corresponding ground-truth color $C_{gt}$:
\begin{align}
\mathcal{L}_\text{rgb} = \rho(\| C - C_{gt} \|)
\end{align}
In addition, we assume an estimate of the human mask available and apply a loss on the rendered 2D alpha values, in order to reduce floating artifacts in space. 
\begin{align}
\mathcal{L}_\text{alpha}  = \| \alpha - \alpha_{gt} \|_1
\end{align}

\paragraph{Hard Surface Regularization} Following \cite{rebain2022lolnerf}, we add further regularization to encourage the NeRF model to predict solid surfaces:
\begin{align}
\mathcal{L}_\text{hard}  = -\log(\exp^{-|\alpha|} + \exp^{-|\alpha-1|}) + const.
\end{align}
where $const.$ is a constant to ensure loss value to be non-negative.
Encouraging solid surfaces helps to speed up rendering because we can terminate rays early once the accumulated opacity reaches 1.

\input{table/snapshot-quantitative.tex}

\paragraph{Occupancy-based regularization} Previous methods for the learning of human avatars \cite{jiang2022neuman, chen2021animatable} often encourage models to predict zero density for points outside of the surface and solid density for points inside the surface by leveraging the SMPL body model as regularizer. This is done to reduce artifacts near the body surface. However such regularization makes heavy assumptions about the shape of the body and does not generalize well for loose clothing. Moreover, we empirically found this regularization is not effective in removing artifacts near the body. This can be seen in Fig.~\ref{fig:peoplsnpshot-qualitative}. Instead of using SMPL for regularization, we use our occupancy grid which is a more conservative estimate of the shape of the subject and the clothing, and define an additional loss $\mathcal{L}_\text{reg}$ which encourages the points inside the empty cells of the occupancy grid to have zero density:
\begin{equation}
\mathcal{L}_\text{density}  = \begin{cases}
    |\sigma(\point)| & \text{if $\point$ is in the empty space}\\
    0 & \text{otherwise}\\
\end{cases}    
\end{equation}

%% file: table/snapshot-quantitative.tex
\begin{table*}[t]
\centering
\definecolor{Gray}{gray}{0.85}
\resizebox{\textwidth}{!}{
\begin{tabular}{rcccccccccccc}
\toprule
& \multicolumn{3}{c}{male-3-casual} & \multicolumn{3}{c}{male-4-casual} & \multicolumn{3}{c}{female-3-casual} & \multicolumn{3}{c}{female-4-casual}\\
& PSNR$\uparrow$  & SSIM$\uparrow$  & LPIPS$\downarrow$  & PSNR$\uparrow$  & SSIM$\uparrow$  & LPIPS$\downarrow$  & PSNR$\uparrow$  & SSIM$\uparrow$  & LPIPS$\downarrow$  & PSNR$\uparrow$  & SSIM$\uparrow$  & LPIPS$\downarrow$ \\
\midrule
Neural Body~\cite{Peng2021CVPR} ($\sim$ 14 hours) & 24.94 & 0.9428 & 0.0326 & 24.71 & 0.9469 & 0.0423& 23.87& 0.9504& 0.0346& 24.37& 0.9451 & 0.0382 \\
Anim-NeRF~\cite{chen2021animatable} ($\sim$ 13 hours) &29.37 & 0.9703 & \textbf{0.0168} & \textbf{28.37} & 0.9605 & \textbf{0.0268} & \textbf{28.91} & \textbf{0.9743} & \textbf{0.0215} &28.90 & 0.9678 & \textbf{0.0174} \\
Ours (1 minute) & \textbf{29.65} & \textbf{0.9730} & 0.0192 & 27.97 & \textbf{0.9649} & 0.0346  & 27.90 & 0.9722 & 0.0249 & \textbf{28.92} & \textbf{0.9692} & 0.0180  \\
\midrule
Anim-NeRF~\cite{chen2021animatable} (5 minutes) &23.17 &0.9266 &0.0784 &22.30 &0.9235 &0.0911 &22.37 &0.9311 &0.0784 &23.18 &0.9292 &0.0687 \\
Ours (5 minutes)  &\textbf{29.53} &\textbf{0.9716} &\textbf{0.0155} &\textbf{27.67} &\textbf{0.9626} &\textbf{0.0307} &\textbf{27.66} &\textbf{0.9709} &\textbf{0.0210} &\textbf{29.11} &\textbf{0.9683} &\textbf{0.0167}\\
\midrule
Anim-NeRF~\cite{chen2021animatable} (3 minutes) & 19.75 & 0.8927 & 0.1286 & 20.66 & 0.8986 & 0.1414 & 19.77 & 0.9003 & 0.1255 & 20.20 & 0.9044 & 0.1109 \\
Ours (3 minutes) &\textbf{29.58} &\textbf{0.9719} &\textbf{0.0157}  &\textbf{27.83} &\textbf{0.9640} &\textbf{0.0342} &\textbf{27.68} &\textbf{0.9708} &\textbf{0.0217} &\textbf{29.05} &\textbf{0.9689} & \textbf{0.0263}\\
\midrule
Anim-NeRF~\cite{chen2021animatable} (1 minute) & 12.39 & 0.7929 & 0.3393 & 13.10 & 0.7705 & 0.3460 & 11.71 & 0.7797 & 0.3321 & 12.31 & 0.8089 & 0.3344\\
Ours (1 minute) & \textbf{29.65} & \textbf{0.9730} & \textbf{0.0192} & \textbf{27.97} & \textbf{0.9649} & \textbf{0.0346}  & \textbf{27.90} & \textbf{0.9722} & \textbf{0.0249} & \textbf{28.92} & \textbf{0.9692} & \textbf{0.0180}  \\

\bottomrule

\end{tabular}
}
\caption{ \textbf{Qualitative Comparison with SoTA on the PeopleSnapshot~\cite{Alldieck20183DV} dataset.} We report PSNR, SSIM and LPIPS~\cite{zhang2018perceptual} between real images and the images generated by our method and two SoTA methods, Neural Body~\cite{Peng2021CVPR} and Anim-NeRF~\cite{chen2021animatable}. We compare all three methods at their convergence, and also compare ours with Anim-NeRF at 5 minutes, 3 minutes and 1 minute training time. }
\label{tab:peoplesnapshot}
\end{table*}

%% file: sec/05-experiment.tex
\section{Experiments}
\label{sec:experiment}

We evaluate the accuracy and speed of our method on monocular videos and compare it with other SoTA methods. In addition, we provide an ablation study to investigate the effect of individual technical contributions.

\subsection{Evaluation Setting}

\subsubsection*{\underline{Datasets}}

\paragraph{PeopleSnapshot} We conduct experiments on the PeopleSnapshot~\cite{Alldieck20183DV} dataset, which contains videos of humans rotating in front of a camera. We follow the evaluation protocol defined in Anim-NeRF~\cite{chen2021animatable}. Because the pose parameters provided in this dataset are not perfect and do not always align with the image, Anim-NeRF optimizes the poses of training and test frames. We train our model with the pose parameters optimized by Anim-NeRF and keep them frozen throughout training for a fair comparison. 

\paragraph{SURREAL} The PeopleSnapshot dataset has limited pose variations. To evaluate the performance on more challenging test poses, we also generate synthetic monocular sequences by rendering SMPL with texture maps from the SURREAL~\cite{varol17_surreal} dataset. For training, we drive the textured SMPL model with the same SMPL parameters from PeopleSnapshot, and for test, we generate challenging out-of-distribution poses. This allows us to evaluate the performance of methods on novel pose synthesis.

\subsubsection*{\underline{Baselines}}
We consider the following methods as our baselines:
\paragraph{Anim-NeRF~\cite{chen2021animatable}} This baseline models human shapes and appearance in a canonical space with an MLP-based NeRF. Given a pose, they first generate a SMPL body in the target pose. Then for each query point in deformed space, its corresponding skinning weights are defined as the weighted average of skinning weights of its K nearest vertices on the posed SMPL mesh. Finally, with the skinning weights, the query point can be transformed back to the canonical space based on inverse LBS.

\paragraph{Neural Body~\cite{Peng2021CVPR}} This baseline learns a set of latent codes anchored to a deformable SMPL mesh. These latent codes deform with the SMPL mesh and are decoded into radiance fields in different poses.

\input{fig/snapshot_images}

\input{fig/training_progression}

\subsection{Comparison with SoTA}

\input{table/synthetic-quantitative.tex}

\input{table/ablation}

\paragraph{Reconstruction Quality} To measure the appearance quality of the reconstructed avatar, we animate and render the reconstructed model with the poses of test frames in PeopleSnapshot, and measure the difference between the generated images and the real images. When training all methods to convergence, our generated images are significantly better than Neural Body~\cite{Peng2021CVPR} and achieve on-par quality as SoTA method Anim-NeRF~\cite{chen2021animatable}, as indicated by the image quality metrics in Tab.~\ref{tab:peoplesnapshot} and the qualitative results in Fig.~\ref{fig:peoplsnpshot-qualitative}.

\paragraph{Speed} Our method requires much less training time and computation resources than SoTA methods. We only require 1 minute training time on a single RTX 3090 while Anim-NeRF~\cite{chen2021animatable} requires 13 hours on $2\times$ RTX 3090 and Neural Body~\cite{Peng2021CVPR} requires 14 hours on $4\times$ RTX 2080. Our method also achieves superior rendering speed - we can render images at $540\times540$ resolution on a single RTX 3090 at 15 FPS, which is orders of magnitude faster than baselines.

Given the same training time budget, our method achieves significantly better image quality than Anim-NeRF as shown in Tab.~\ref{tab:peoplesnapshot}. Comparing our training progression with Anim-NeRF in Fig.~\ref{fig:progression}, we note that our method already learns meaningful appearance and moderate details within 5s and acceptable visual quality at 10s. After only 1 minute of training time, our method already achieves high-fidelity reconstruction quality. In contrast, Anim-NeRF does not produce meaningful results this early in training and only learns the rough shape after 3 minutes.

\paragraph{Novel Pose Synthesis Quality}
The previous evaluation does not reflect the performance of novel pose synthesis, because the pose variation in the PeopleSnapshot dataset is limited (self-rotating). Due to the lack of ground truth images in novel poses, we resort to evaluating novel pose synthesis qualitatively. We generate images in novel challenging poses with our method and Anim-NeRF. As shown in Fig.~\ref{fig:peoplsnpshot-qualitative}, our method can faithfully generate images even in challenging body poses while preserving high fidelity. In contrast, Anim-NeRF suffers from artifacts under arms and between legs, because their methods cannot correctly disambiguate body parts that are close to each other in the posed space. Our method outperforms our baseline especially for loose clothing as shown in the bottom example in Fig.~\ref{fig:peoplsnpshot-qualitative}. This is because we don't rely on the SMPL body model for regularization and hence can better deal with subjects and clothing that differ from SMPL. To quantitatively evaluate novel pose synthesis, we generate synthetic data in challenging poses as ground-truth. The results in Tab.~\ref{tab:synthetic} and Fig.~\ref{fig:peoplsnpshot-qualitative} verify the superiority of our method in terms of novel pose synthesis quality.

\input{fig/ablation/occupancy_cache/occupancy_cache.tex}

\input{fig/more_results.tex}

\subsection{Ablation Study}

\paragraph{Empty Space Skipping} We study the effect of our proposed empty space skipping scheme for dynamic objects. As shown in Tab.~\ref{tab:abltion_speed}, skipping empty space significantly improves the training and rendering speed.

\paragraph{Occupancy-based Regularization $\loss_{reg}$} The occupancy grid for empty space skipping can also help regularize the radiance field to reduce noise via our regularization loss $\loss_{reg}$ described in Section.~\ref{sec:method:losses}. As shown in Fig.~\ref{fig:reg}, this loss effectively reduces floating artifacts and consequently helps to improve the overall image quality as evidenced by the PSNR improvement in Tab.~\ref{tab:abltion_reg}. Another common approach to reducing floating noise is to encourage zero density for every point in space. We compare our solution with this strategy and find that this strategy (Global Sparsity) leads to degenerated image quality as shown in Fig.~\ref{fig:reg}.

\input{sec/06-ablation.tex}
\subsection{Additional Qualitative Samples}
We show additional qualitative samples in Fig.~\ref{fig:more_snapshot}.

%% file: fig/snapshot_images.tex
\begin{figure*}[t]
\begin{center}
\setlength\tabcolsep{1pt}
\newcommand{\crop}{0.8cm}
\newcommand{\cropsmall}{0.4cm}
\newcommand{\height}{2.8cm}
\begin{tabularx}{\linewidth}{ l cccccc }
& Input & View 1 & View 2 & Pose 1 & Pose 2 & Pose 3\\
\rotatebox{90} {~~~~~Anim-NeRF~~~~~} 

    &  \includegraphics[height=\height, width=\height, clip]{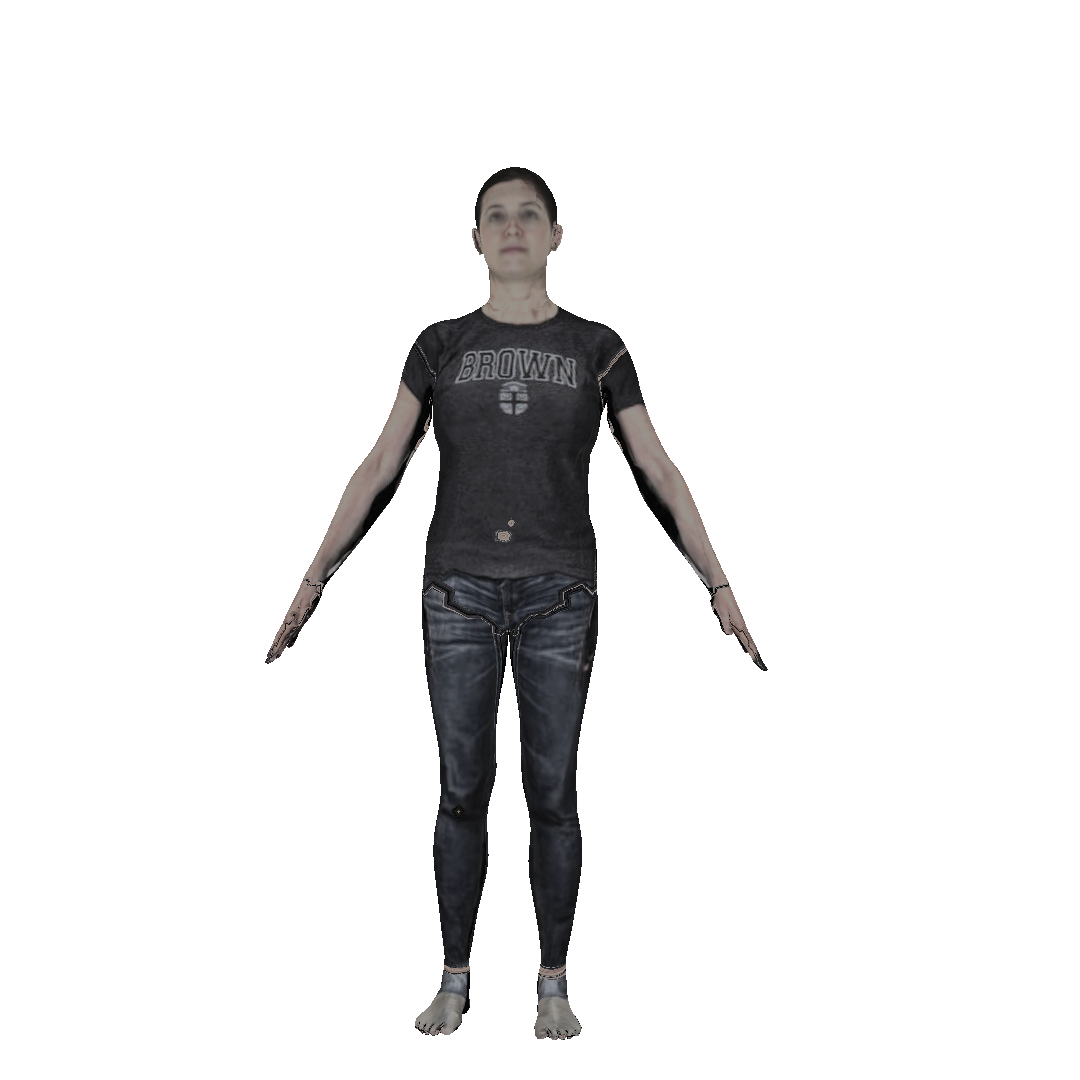}
    &  \includegraphics[height=\height, width=\height, clip]{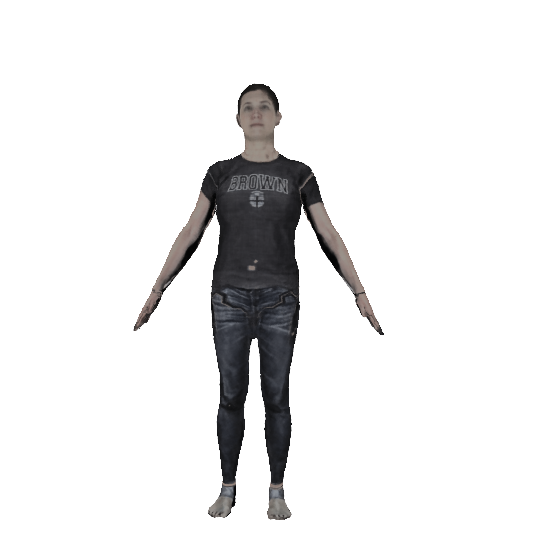}
    &  \includegraphics[height=\height, width=\height, clip]{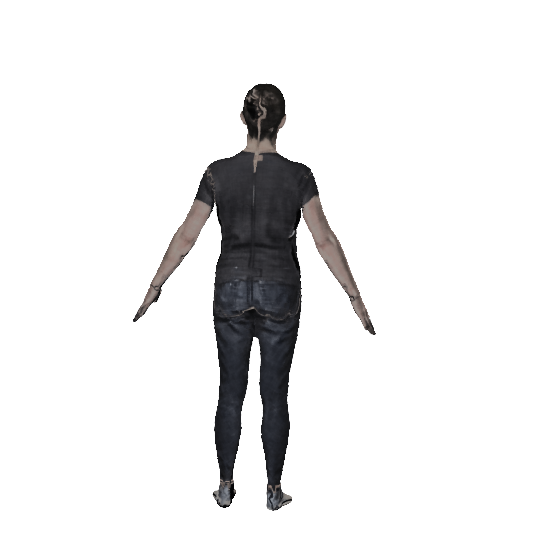}
    &  \includegraphics[height=\height, width=\height, clip]{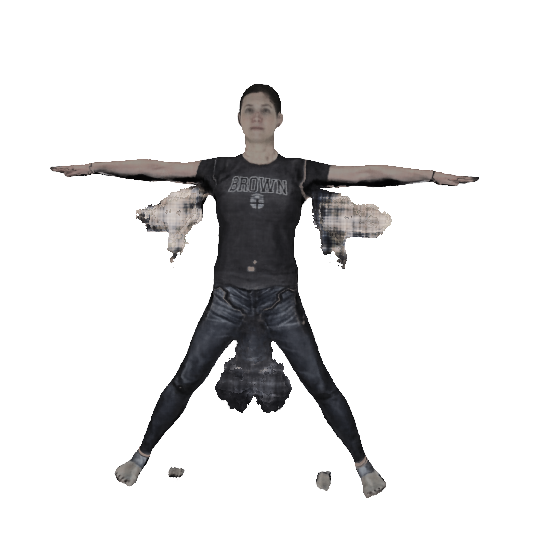}
    &  \includegraphics[height=\height, width=\height, clip]{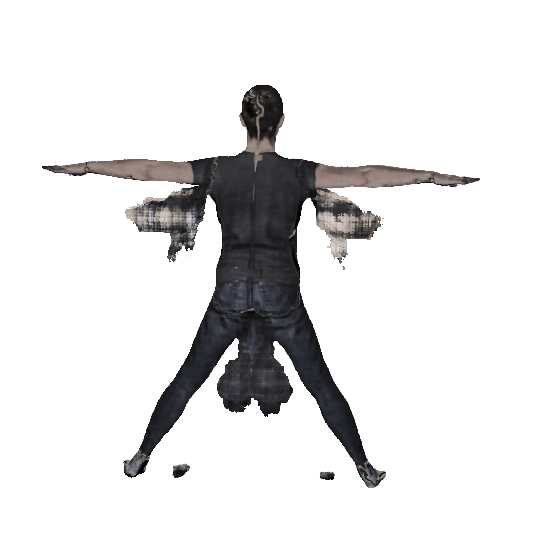}
    &  \includegraphics[height=\height, width=\height, clip]{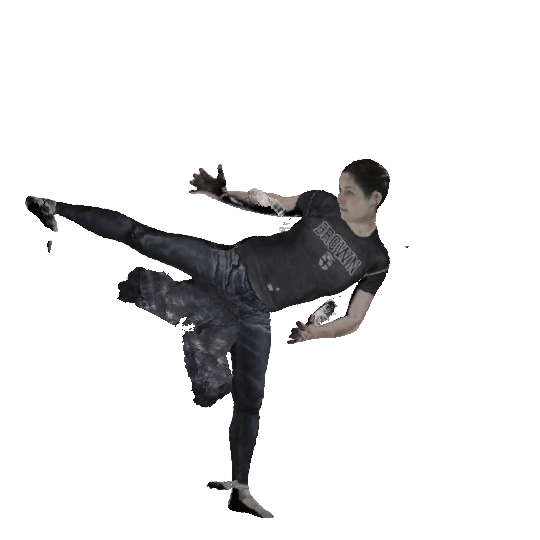}
    \\
    \rotatebox{90} {~~~~~~~~~Ours~~~~~}
    &  \includegraphics[height=\height]{fig/synthetic/ours/S3/input.png}
    &  \includegraphics[height=\height, width=\height, clip]{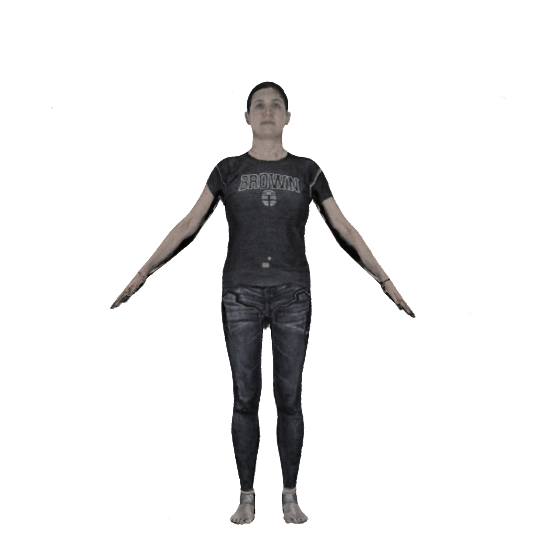}
    &  \includegraphics[height=\height, width=\height, clip]{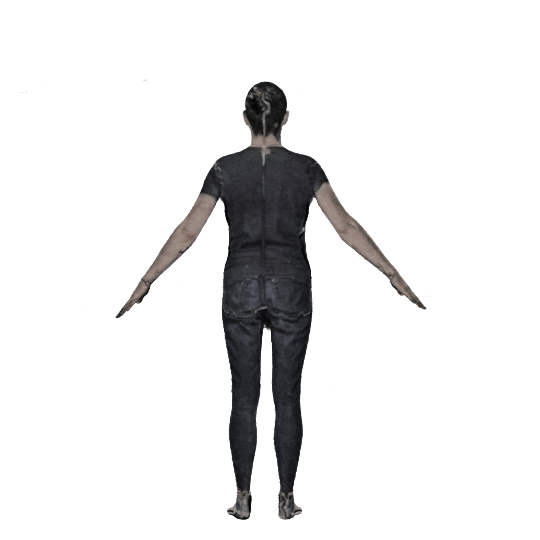}
    &  \includegraphics[height=\height, width=\height, clip]{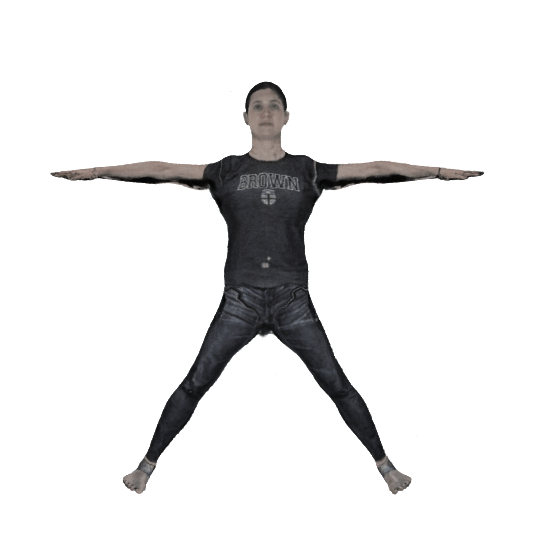}
    &  \includegraphics[height=\height, width=\height, clip]{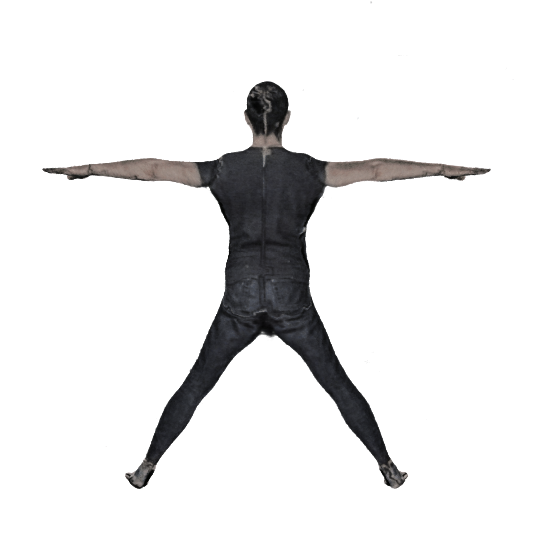}
    &  \includegraphics[height=\height, width=\height, clip]{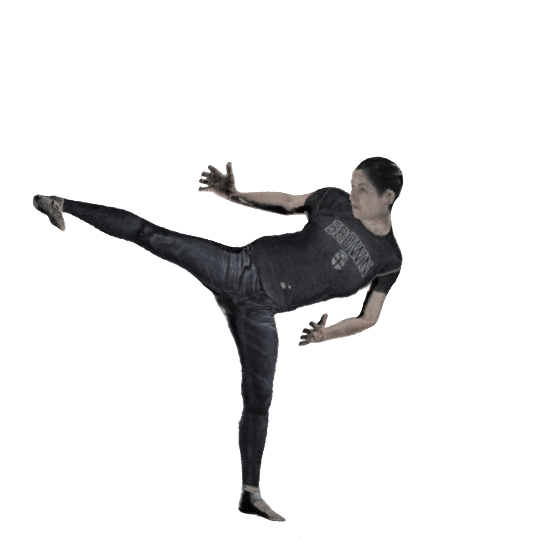}
    \\
    
\rotatebox{90} {~~~~~Anim-NeRF~~~~~} 
    &  \includegraphics[height=\height]{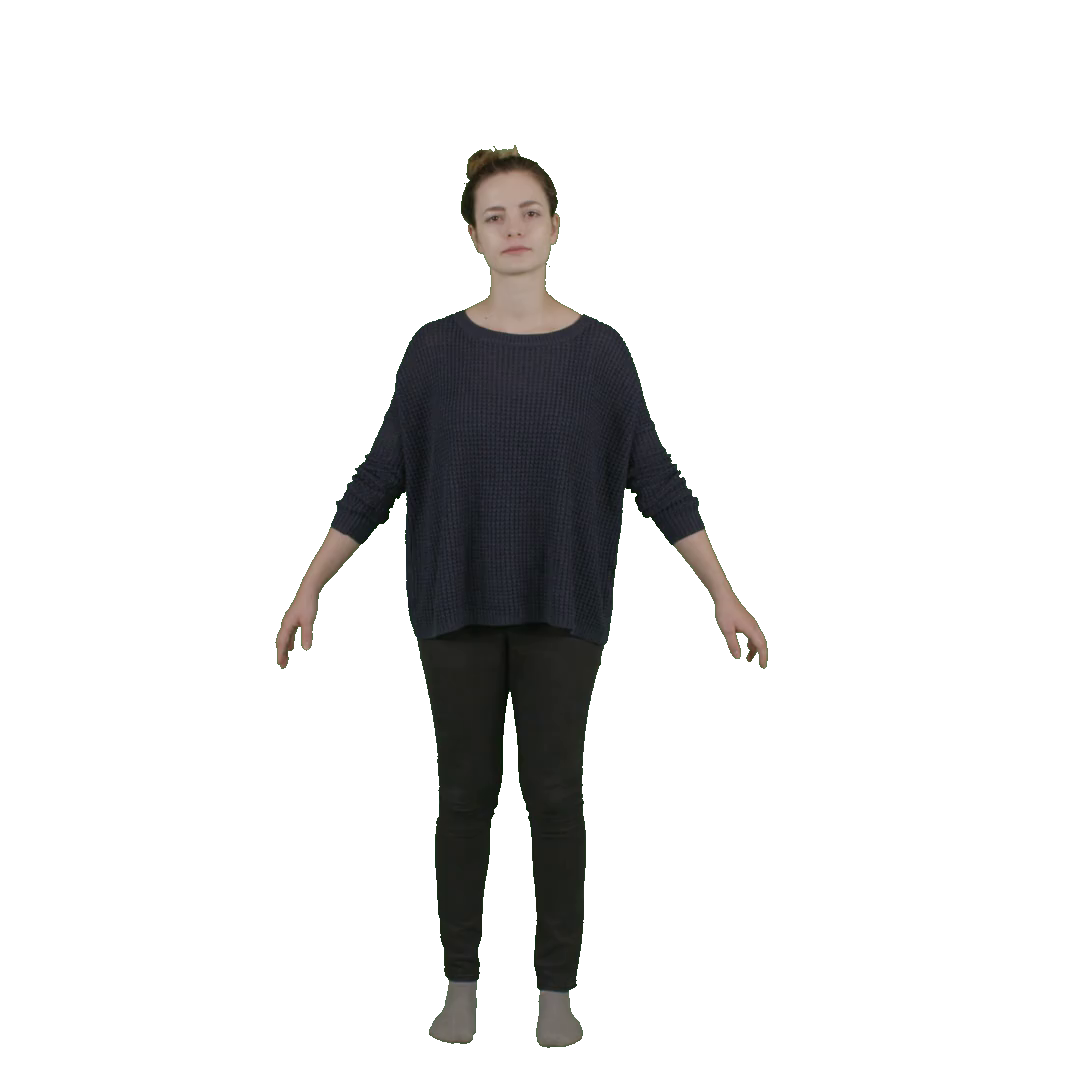} 
    &  \includegraphics[height=\height, width=\height, clip]{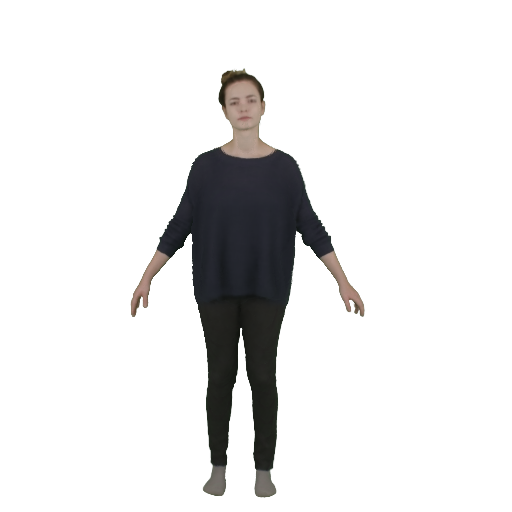}
    &  \includegraphics[height=\height, width=\height, clip]{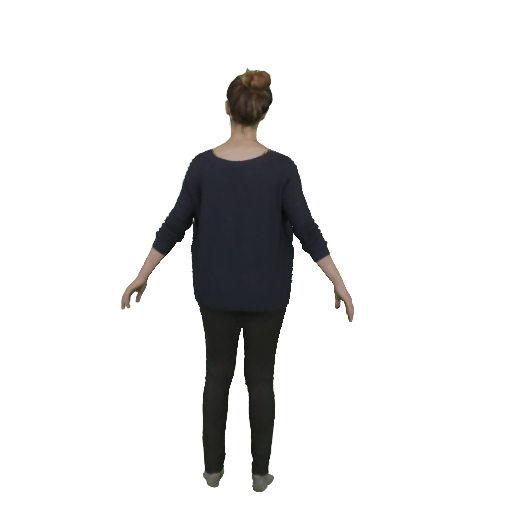}
    &  \includegraphics[height=\height, width=\height, clip]{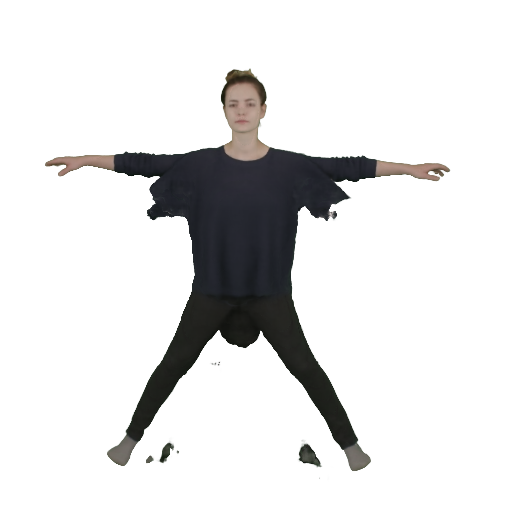}
    &  \includegraphics[height=\height, width=\height, clip]{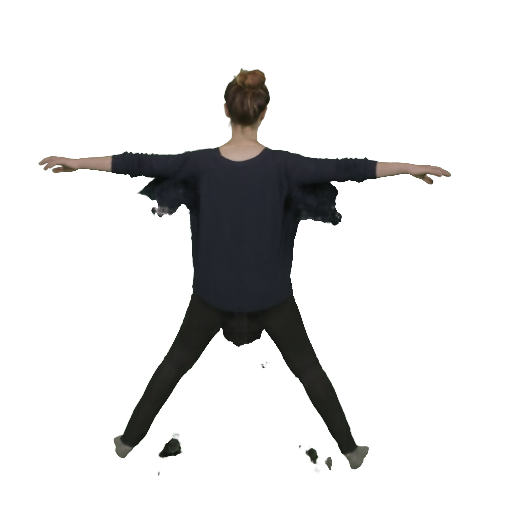}
    &  \includegraphics[height=\height, width=\height, clip]{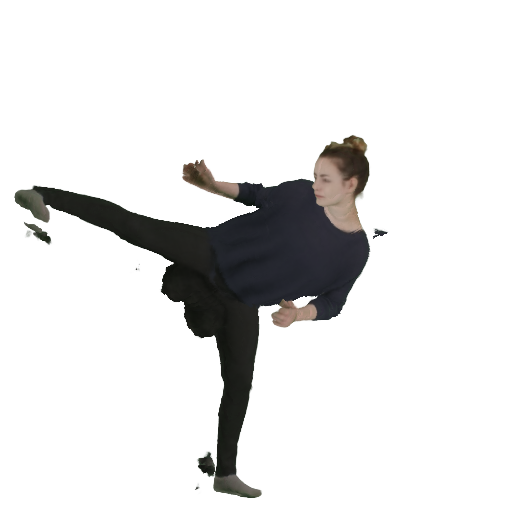}
    \\
    \rotatebox{90} {~~~~~~~~~Ours~~~~~}
    &  \includegraphics[height=\height]{fig/peoplesnapshot/qualitative/f3c/gt.png} 
    &  \includegraphics[height=\height]{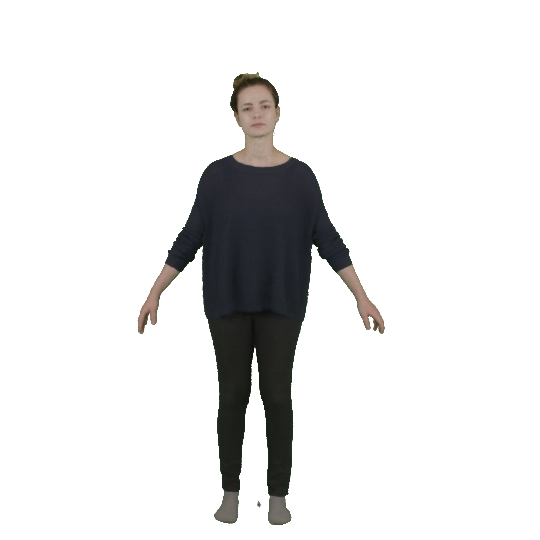} 
    &  \includegraphics[height=\height]{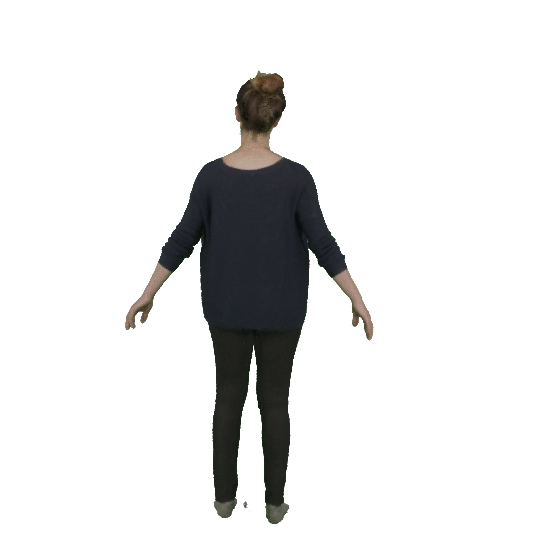} 
    &  \includegraphics[height=\height, width=\height, clip]{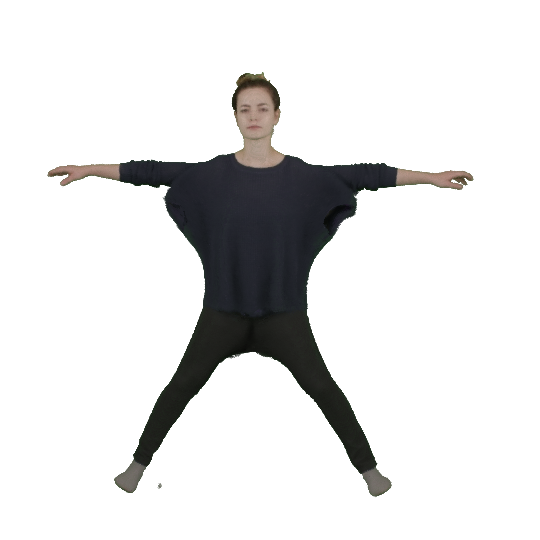}
    &  \includegraphics[height=\height, width=\height, clip]{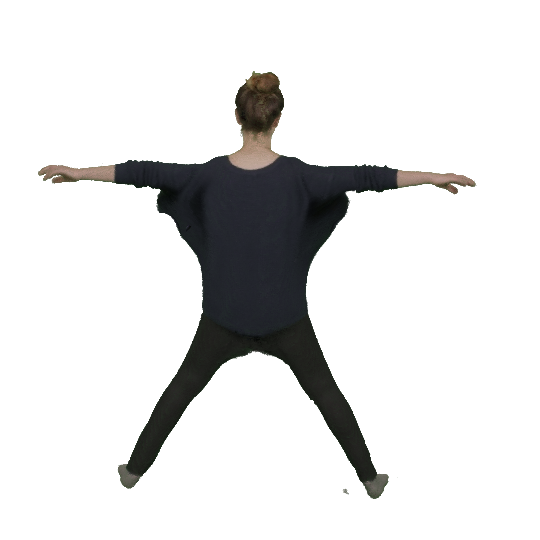}
    &  \includegraphics[height=\height, width=\height, clip]{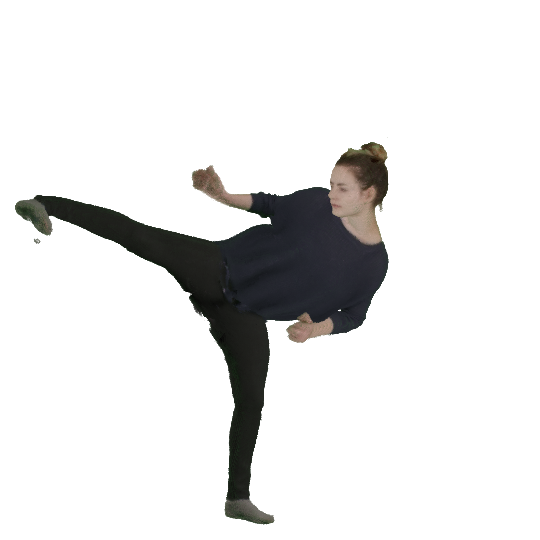}
    \\

\end{tabularx}
\vspace{-0.5em}
\caption{\textbf{Qualitative Results on SURREAL~\cite{varol17_surreal} and PeopleSnapshot dataset~\cite{Alldieck20183DV}.} We show reconstructed avatars on SURREAL (top) and PeopleSnapshot (bottom) from different viewpoints (column 2-3) and in various poses (column 4-6).
}
\vspace{-1.5em}
\label{fig:peoplsnpshot-qualitative}
\end{center}
\end{figure*}

%% file: fig/training_progression.tex
\begin{figure*}[t!]
\begin{center}
\setlength\tabcolsep{1pt}
\newcommand{\crop}{0}
\newcommand{\height}{3.3cm}
\begin{tabularx}{\linewidth}{ l ccccc }
& 5s & 10s & 1 min & 3 min & 5 min\\
\rotatebox{90} {~~~~~Anim-NeRF~~~~~} 
    &  \includegraphics[width=\height, trim={\crop} 0 {\crop} 0, clip]{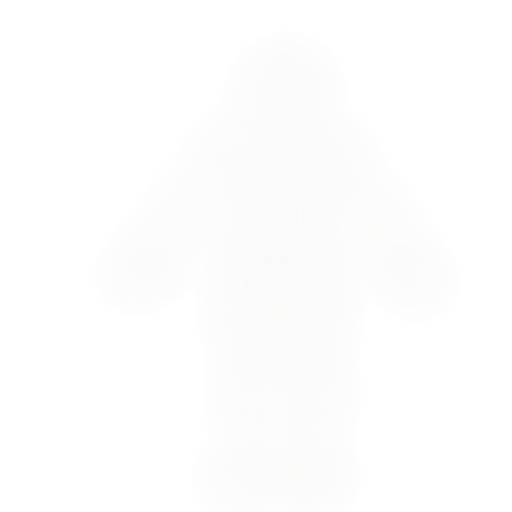}
    &  \includegraphics[width=\height, trim={\crop} 0 {\crop} 0, clip]{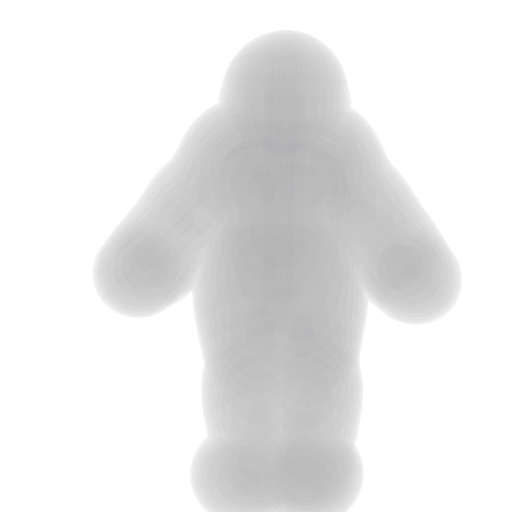}
    &  \includegraphics[width=\height, trim={\crop} 0 {\crop} 0, clip]{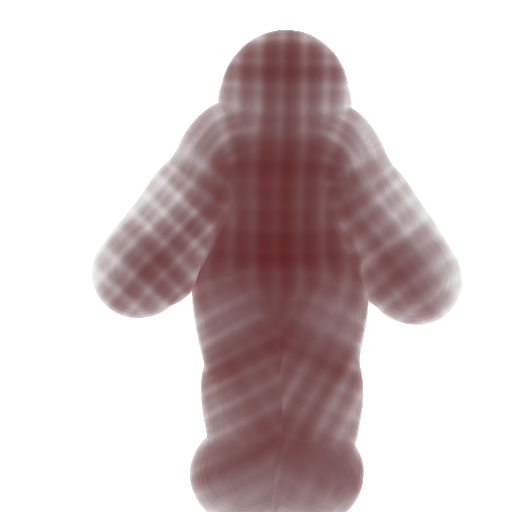}
    &  \includegraphics[width=\height, trim={\crop} 0 {\crop} 0, clip]{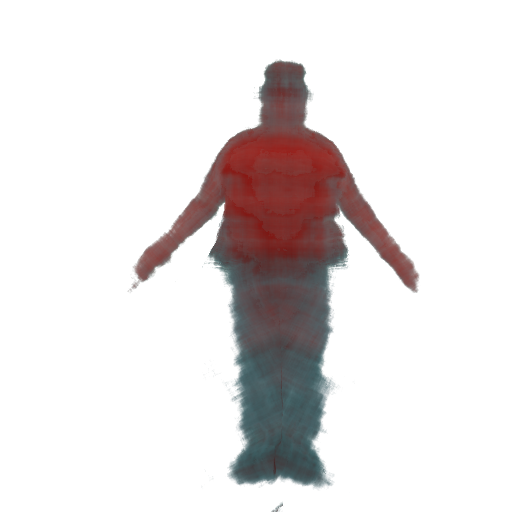}
    &  \includegraphics[width=\height, trim={\crop} 0 {\crop} 0, clip]{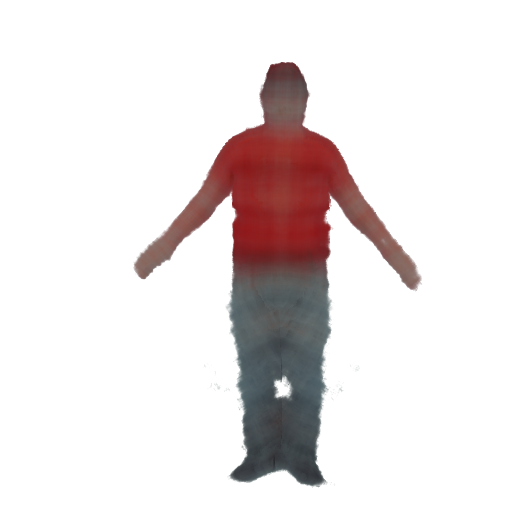}\\
\rotatebox{90} {~~~~~~~~~~~~~Ours~~~~~}
    &  \includegraphics[width=\height, trim={\crop} 0 {\crop} 0, clip]{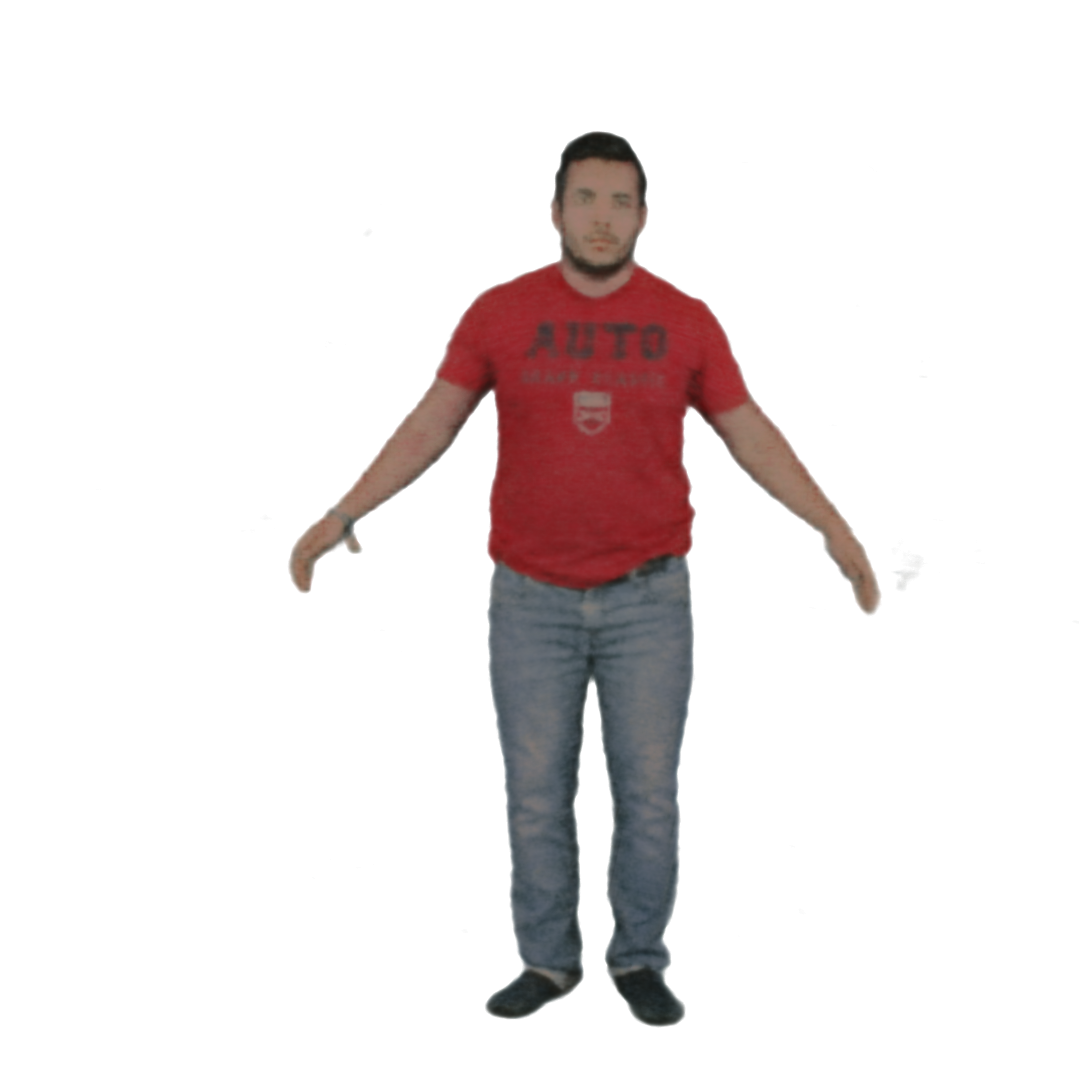}
    &  \includegraphics[width=\height, trim={\crop} 0 {\crop} 0, clip]{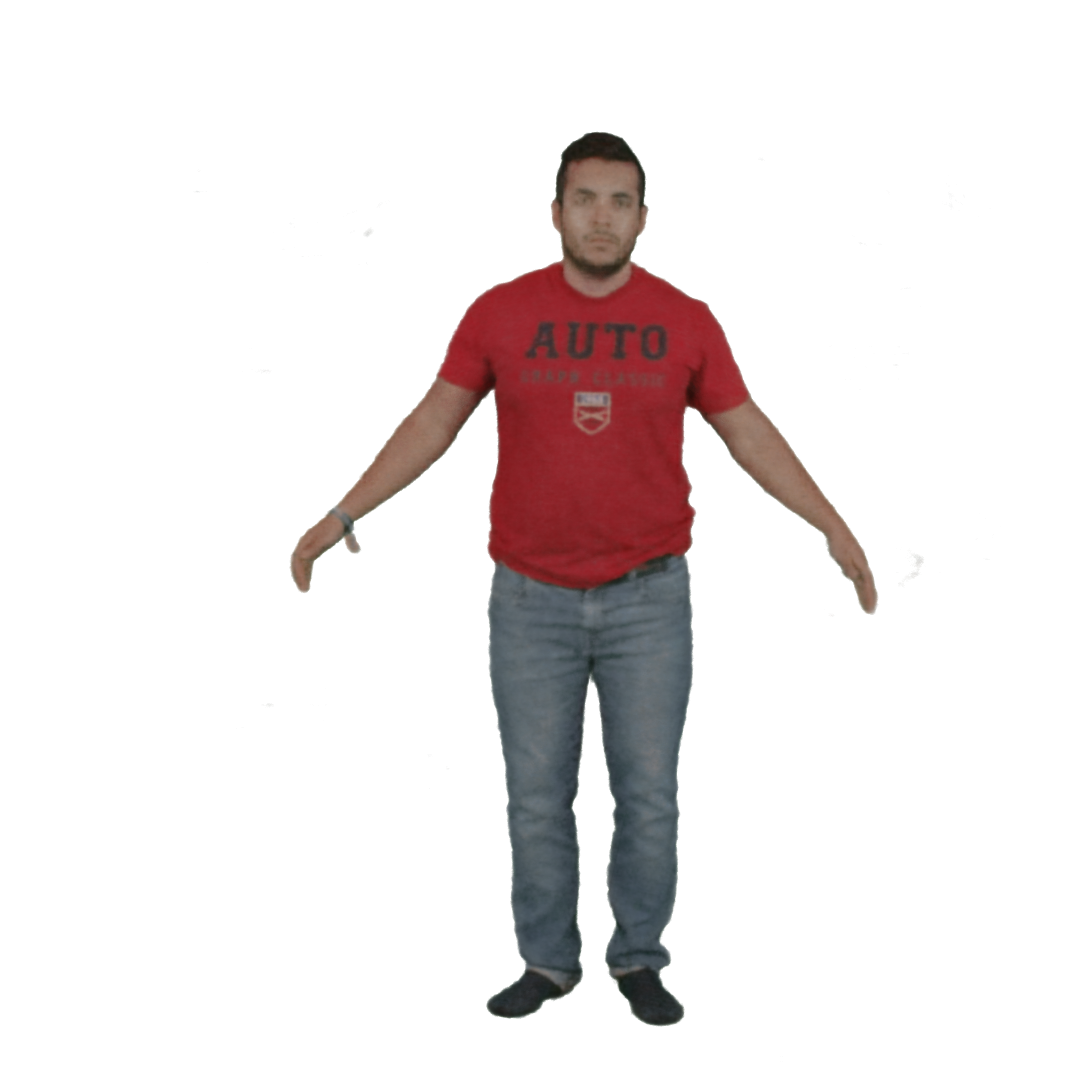}
    &  \includegraphics[width=\height, trim={\crop} 0 {\crop} 0, clip]{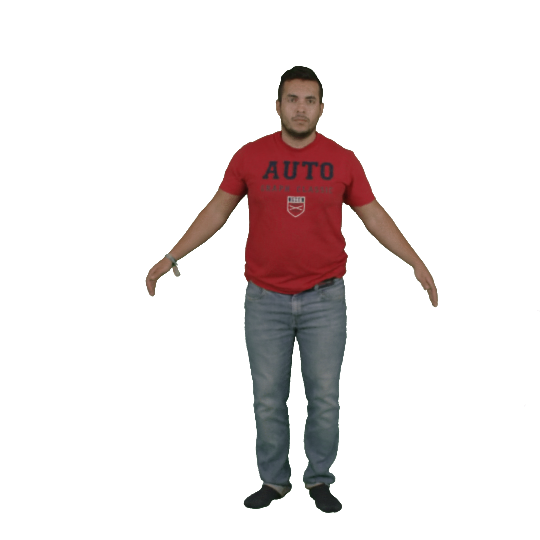}
    &  \includegraphics[width=\height, trim={\crop} 0 {\crop} 0, clip]{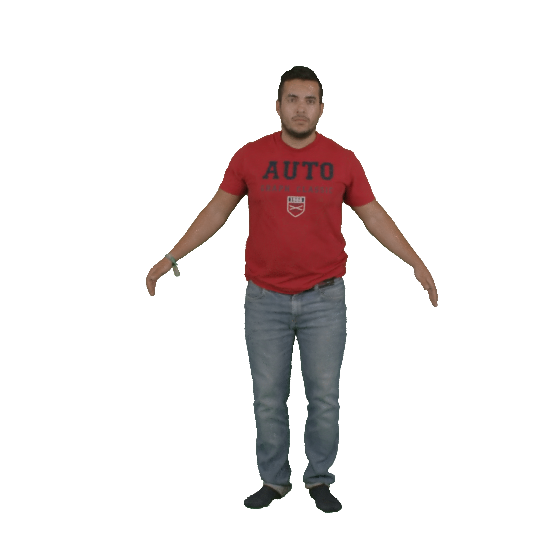}
    &  \includegraphics[width=\height, trim={\crop} 0 {\crop} 0, clip]{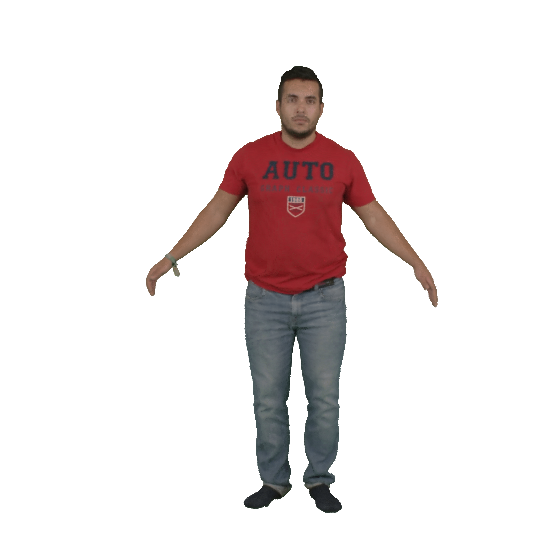}
\end{tabularx}
\vspace{-0.5em}
\caption{\textbf{Training Progression.} We show the image quality at different training iterations. Our method converges significantly faster than SoTA Anim-NeRF~\cite{chen2021animatable}. }
\vspace{-1.5em}
\label{fig:progression}
\end{center}
\end{figure*}

%% file: table/synthetic-quantitative.tex
\begin{table}[tb]
\centering
\resizebox{\linewidth}{!}{
\begin{tabular}{c c c  c c  c c}
\toprule
\multirow{2}{*}{} &  \multicolumn{3}{c}{Anim-NeRF} & \multicolumn{3}{c}{Ours} \\ 
 & PSNR & SSIM & LPIPS & PSNR & SSIM& LPIPS \\ 
\midrule
S1 & 21.66 & \textbf{0.9450} & 0.07615 & \textbf{24.48} & 0.9353 & \textbf{0.0304}\\
S2 & 20.00 & \textbf{0.9483} & 0.09693 & \textbf{23.94} & 0.9354 & \textbf{0.0343} \\
S3 & 20.06 & 0.9326 & 0.07948 & \textbf{25.08} & \textbf{0.9494} & \textbf{0.0275} \\
\bottomrule
\end{tabular}}
\caption{\textbf{Qualitative Results on the SURREAL Dataset.} We evaluate novel pose synthesis quality of our method and Anim-NeRF~\cite{chen2021animatable} on 3 synthetic subjects.}
\label{tab:synthetic}
\end{table}

%% file: table/ablation.tex
\begin{table}[tb]
  \centering
  \begin{tabular}{@{}lcc@{}}
    \toprule
                     & Training & Rendering \\
    \midrule
    w/o empty space skipping & 3m 10s & $\sim$ 1 FPS \\
    w/  empty space skipping & \textbf{1m 47s} & $\sim$ \textbf{15 FPS} \\
    \bottomrule
  \end{tabular}
  \caption{\textbf{Empty Space Skipping.} We compare the training and rendering speed with and without empty space skipping. For the training time we report the average training time of 100 epochs among 4 sequences in PeopleSnapshot. }
  \label{tab:abltion_speed}
\end{table}

\begin{table}[tb!]
  \centering
  \resizebox{\linewidth}{!}{
  \begin{tabular}{@{}lccc@{}}
    \toprule
                     & PSNR & SSIM & LPIPS\\
    \midrule
    w/o occupancy-based regularizer & 28.22 & 0.9680 & 0.0301\\
    w/  occupancy-based regularizer & \textbf{28.64} & \textbf{0.9700} & \textbf{0.0240}\\
    \bottomrule
  \end{tabular}
  }
  \caption{\textbf{Occupancy-based Regularizer.} We evaluate image quality averaged over the 4  PeopleSnapshot sequences. For both cases we train our model for 100 epochs.
  }
  \label{tab:abltion_reg}
\end{table}

%% file: fig/ablation/occupancy_cache/occupancy_cache.tex
 \newcommand{\tvwidth}{0.24\linewidth}

\begin{figure}[tb!]
  \captionsetup[subfigure]{labelformat=empty}
  \centering
    \begin{subfigure}[b]{0.32\linewidth}
        \includegraphics[trim={3cm 0cm 3cm 3cm}, clip, width=\linewidth]{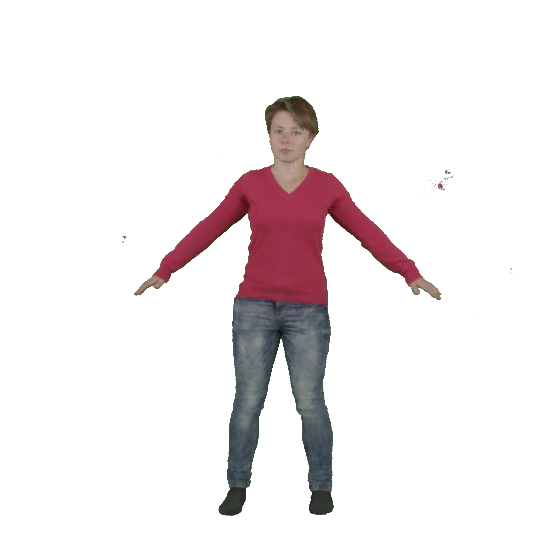}
        \caption{No Reg }
    \end{subfigure}
    \begin{subfigure}[b]{0.32\linewidth}
        \includegraphics[trim={3cm 0cm 3cm 3cm}, clip, width=\linewidth]{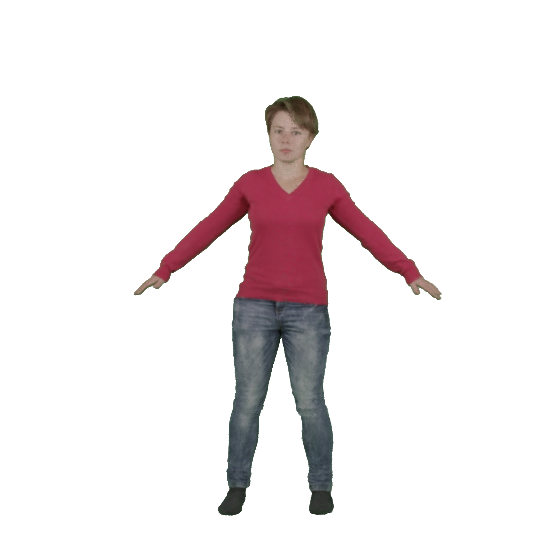}
        \caption{Ours}
    \end{subfigure}
    \begin{subfigure}[b]{0.32\linewidth}
    \includegraphics[trim={3cm 0cm 3cm 3cm}, clip, width=\linewidth]{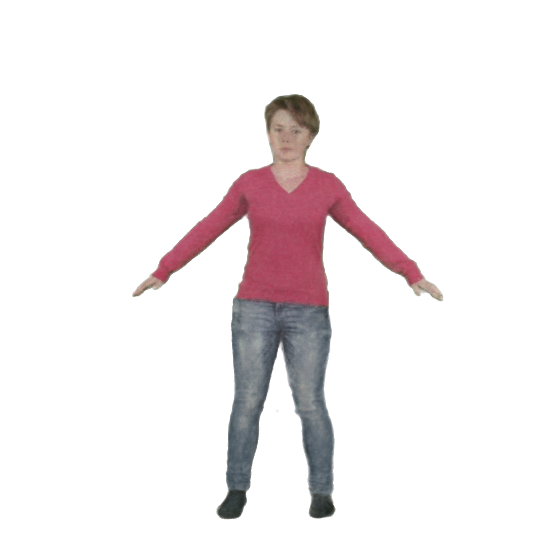}
    \caption{Global Sparsity}
\end{subfigure}
  \caption{\textbf{Effect of Occupancy-based Regularization.} Without our regularization loss, the model suffers from floating artifacts. Our occupancy-based regularization loss successfully removes such artifacts. While a global sparsity prior biasing all densities towards 0 can also reduce such artifacts, it leads to degenerated image quality (semi-transparent).
  }
    \label{fig:reg}
\end{figure}

%% file: fig/more_results.tex
\begin{figure*}[t!]
\centering
\setlength\tabcolsep{0pt}
\newcommand{\cropsmall}{0.4cm}
\newcommand{\height}{3.7cm}
\begin{tabularx}{\linewidth}{ccccc}
\hspace{-2em}Input & View 1 & View 2 & View 3 & Novel Pose \\
\hspace{-2em}\includegraphics[height=\height]{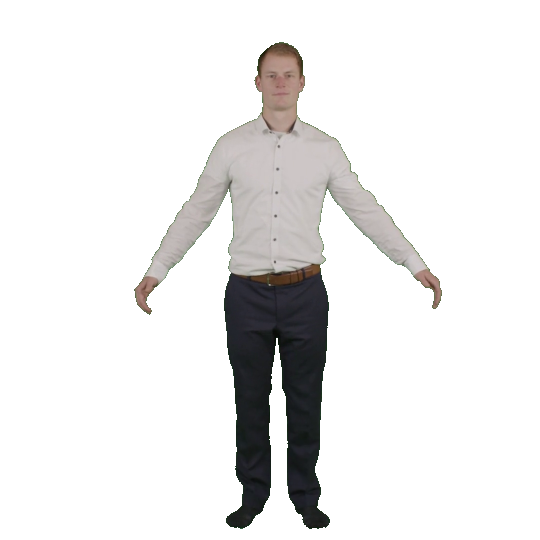} 
    &  \includegraphics[height=\height, width=\height, clip]{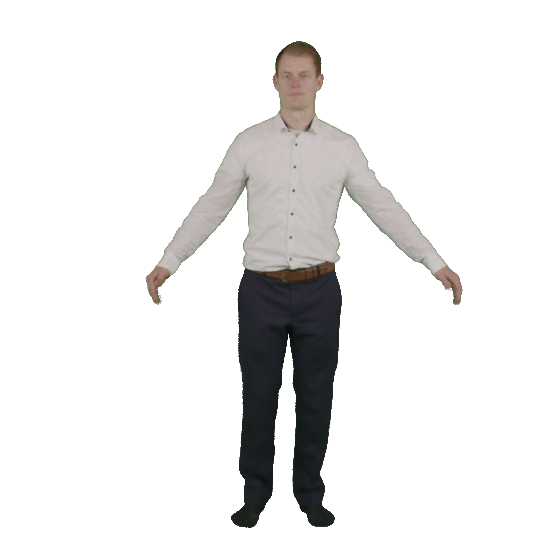} 
    &  \includegraphics[height=\height, width=\height, clip]{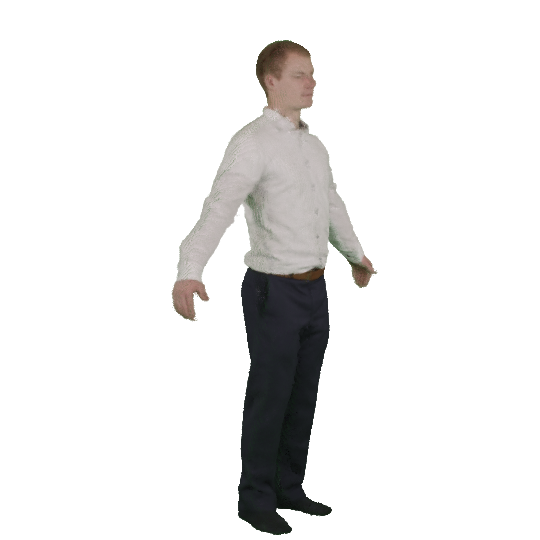} 
    &  \includegraphics[height=\height, width=\height, clip]{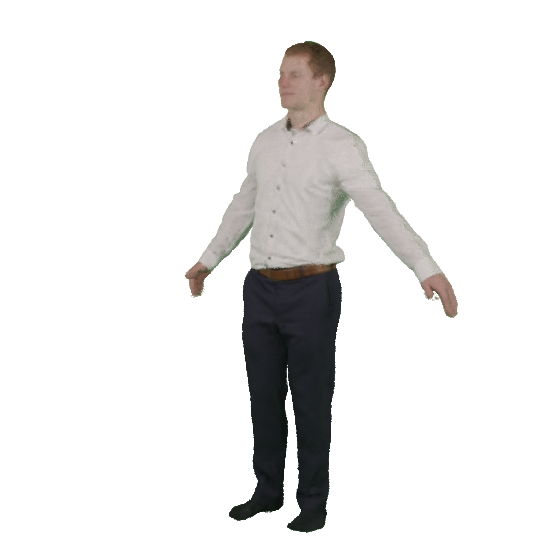} 
    &  \includegraphics[height=\height, width=\height, clip]{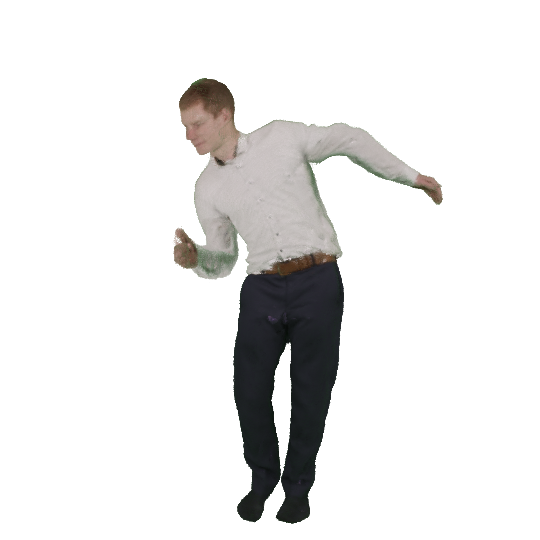}
    \\
\hspace{-2em}\includegraphics[height=\height]{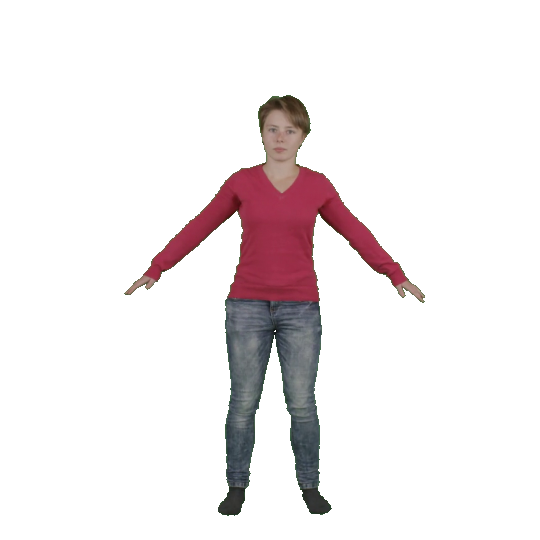} 
    &  \includegraphics[height=\height, width=\height, clip]{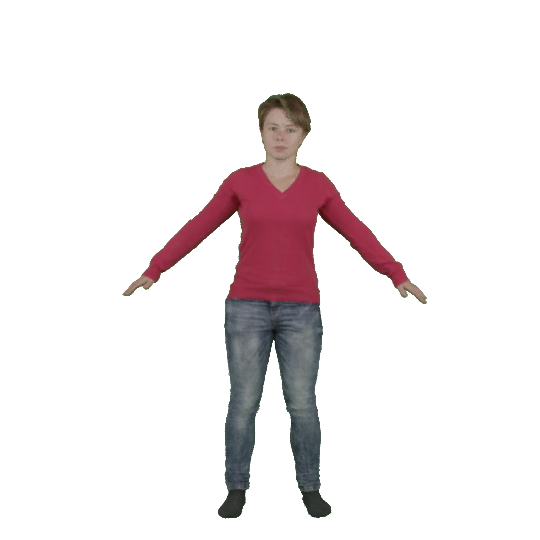} 
    &  \includegraphics[height=\height, width=\height, clip]{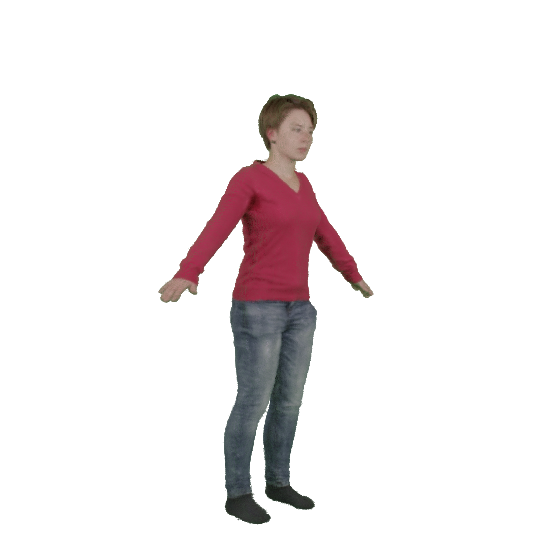} 
    &  \includegraphics[height=\height, width=\height, clip]{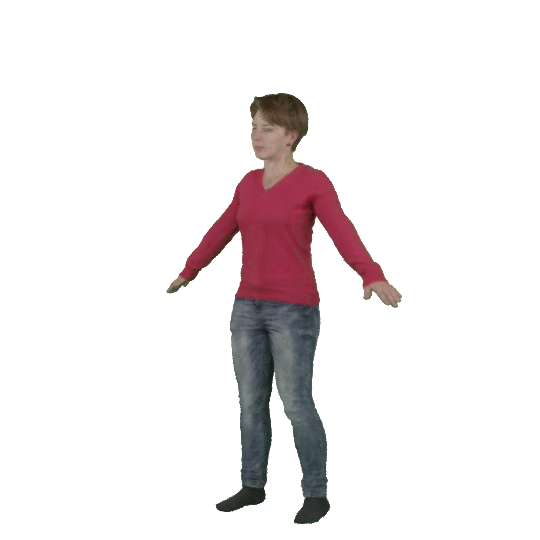} 
    &  \includegraphics[height=\height, width=\height, clip]{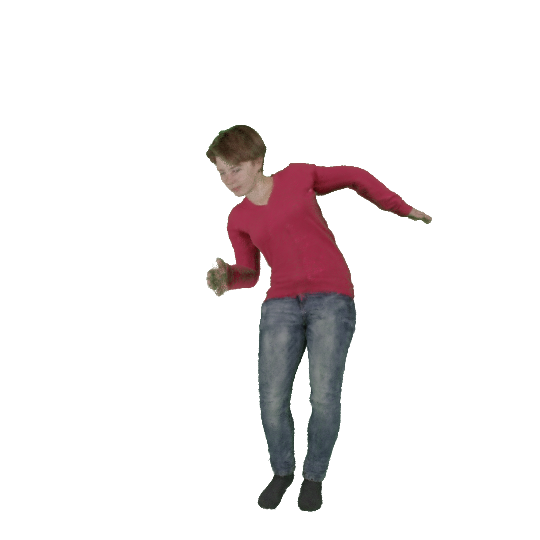}
    \\
\hspace{-2em}\includegraphics[height=\height]{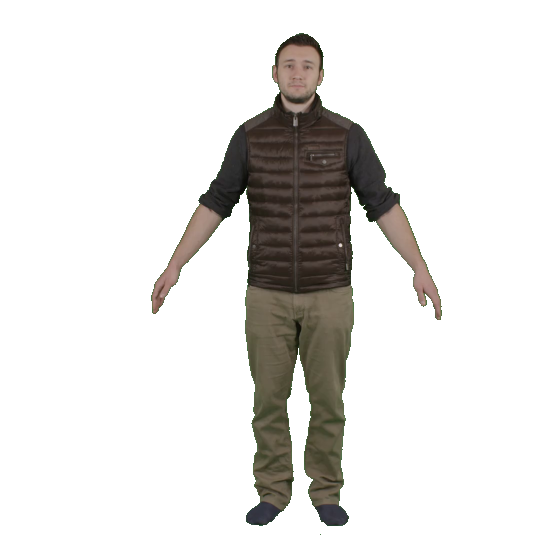} 
    &  \includegraphics[height=\height, width=\height, clip]{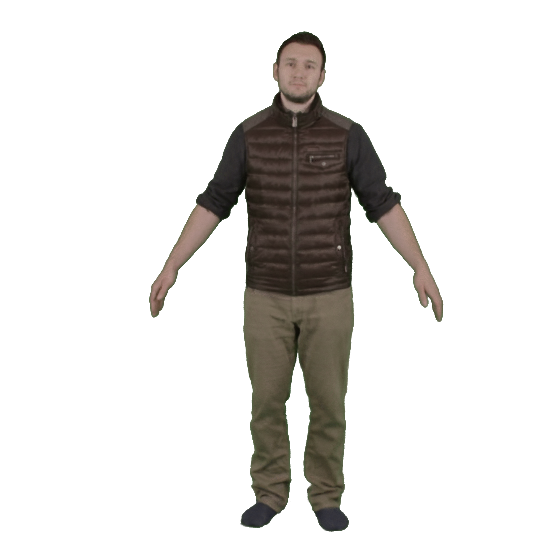} 
    &  \includegraphics[height=\height, width=\height, clip]{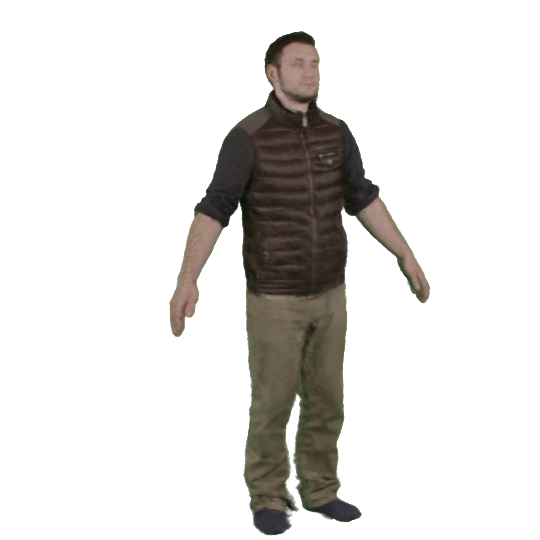}
    &  \includegraphics[height=\height, width=\height, clip]{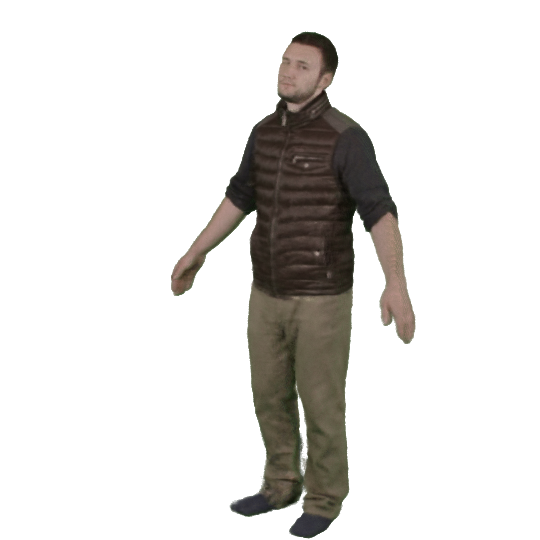} 
    &  \includegraphics[height=\height, width=\height, clip]{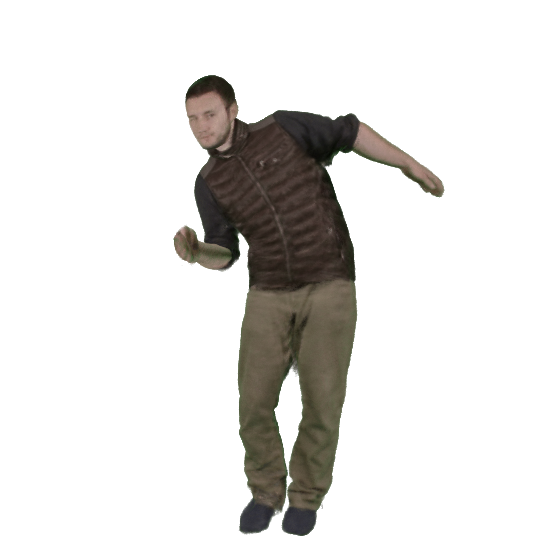}
    \\
\hspace{-2em}\includegraphics[height=\height]{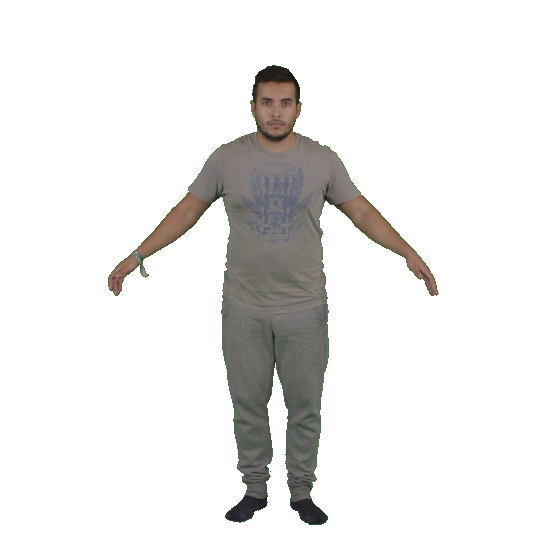} 
    &  \includegraphics[height=\height, width=\height, clip]{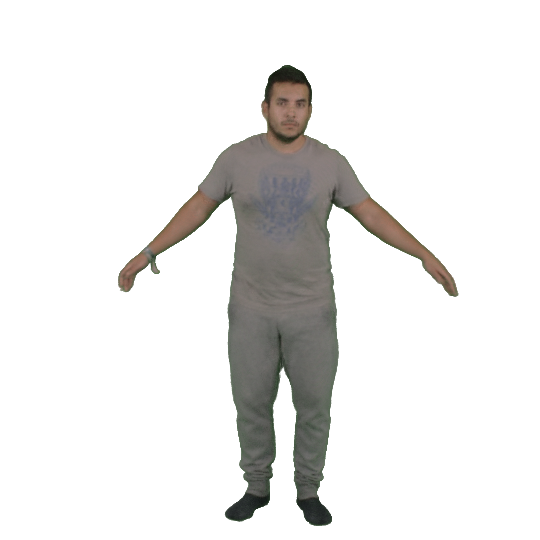} 
    &  \includegraphics[height=\height, width=\height, clip]{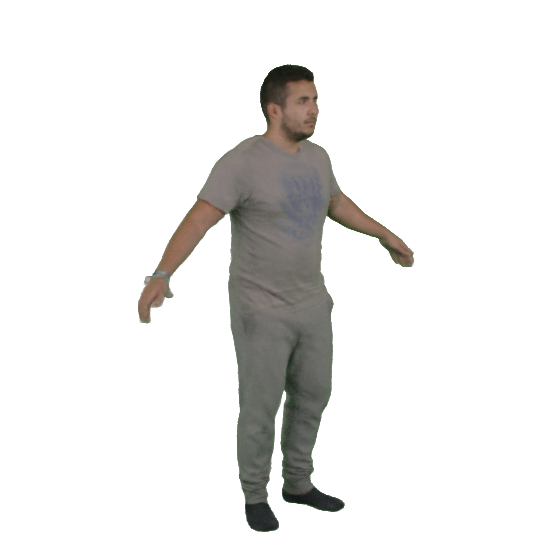} 
    &  \includegraphics[height=\height, width=\height, clip]{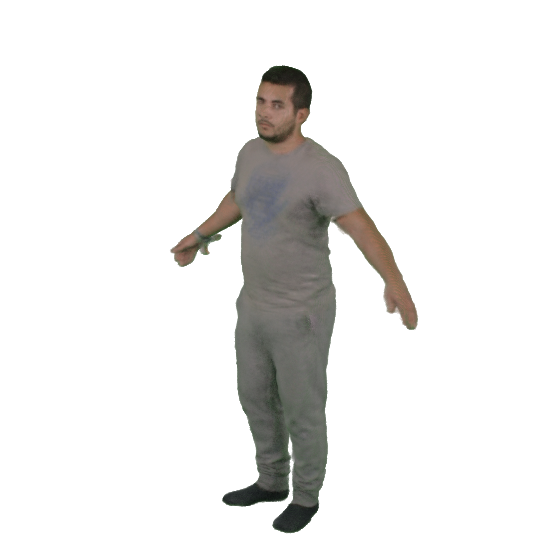} 
    &  \includegraphics[height=\height, width=\height, clip]{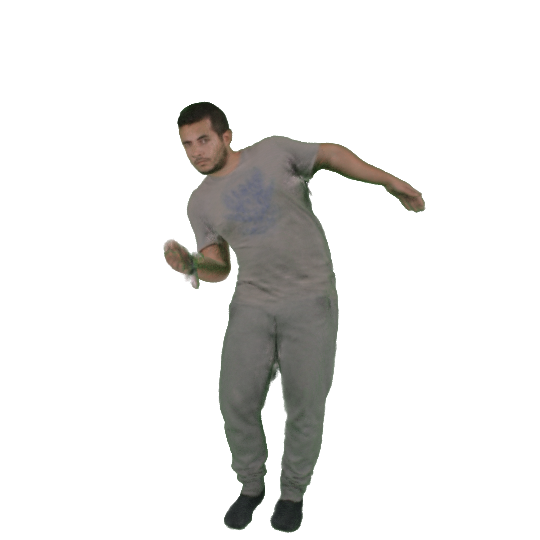}
    \\
\end{tabularx}
\caption{\textbf{More Qualitative Results of Our Method.} 
}
\vspace{-1.5em}
\label{fig:more_snapshot}
\end{figure*}

%% file: sec/07-conclusion.tex
\section{Conclusion}
\label{sec:conclusion}
In this paper, we propose a method that can reconstruct animatable human avatars from monocular videos within 60 seconds and can animate and render the model afterward at 15 FPS. To achieve this, we combine an efficient neural representation, Instant-NGP~\cite{mueller2022instant}, and an efficient articulation module Fast-SNARF~\cite{fastsnarf}. This naive combination does not yield optimal speed. We propose an empty space skipping scheme to improve our rendering speed, and an occupancy-aware regularization loss to reduce floating artifacts in space. In comparison with SoTA methods, our method achieves on-par image quality while being significantly faster during training and inference. 

\paragraph{Limitations and Future Work} 
Our method reconstructs avatars purely based on image observations and therefore cannot infer unseen regions. For instance, if the input video only captures the front side of the subject, our method cannot reconstruct the back side. This limitation could potentially be addressed by leveraging learning-based methods to predict the texture and geometry of unobserved regions. While this paper focuses on full-body human reconstruction, the idea could be applied to other objects. An interesting next step is to extend our method to reconstruct general articulated objects or animals from images efficiently.

\paragraph{Acknowledgements} Xu Chen was supported by the Max Planck ETH Center for Learning Systems. 